\def\withpreprintappendix{1}
\documentclass[journal]{IEEEtran}
\usepackage{amsmath,amsfonts,amssymb}
\usepackage{algorithmic}
\usepackage{algorithm}
\usepackage{array}
\usepackage[caption=false,font=normalsize,labelfont=sf,textfont=sf]{subfig}
\usepackage{textcomp}
\usepackage{stfloats}
\usepackage{url}
\usepackage{verbatim}
\usepackage{graphicx}
\usepackage{cite}
\usepackage{booktabs}
\usepackage{xcolor}
\ifdefined\withpreprintappendix
\else
  \usepackage{xr}                 % cross-reference the separate supplement PDF
  \makeatletter
  \let\sv@bibcite\bibcite
  \let\bibcite\@gobbletwo
  \let\bibcite\sv@bibcite
  \makeatother
\fi
\begin{document}

\title{Separating Decision-Rule Misalignment from Readout-Coverage Limitations in Speech Language Models}

\author{Linkai~Peng~and~Baorian~Nuchged% <-this % stops a space
\thanks{Linkai Peng is with the Institute for the Brain and Cognitive Sciences,
University of Connecticut, Storrs, CT 06269, USA (e-mail: linkai.peng@uconn.edu). Baorian Nuchged is with the Department of Linguistics, The University of Texas at Austin, Austin, TX 78712, USA (e-mail: baorian@utexas.edu).}% <-this % stops a space
% \thanks{Manuscript submitted for review.}}
\thanks{Preprint.}}
% \markboth{IEEE/ACM Transactions on Audio, Speech, and Language Processing}%
% {Peng and Nuchged: Decision-Rule Misalignment and Readout Coverage}

% \IEEEpubid{0000--0000/00\$00.00~\copyright~2026 IEEE}
\maketitle

% SLM on paralingual -> behavioral answer -> cannot seperate failures
% our methods: two gaps; one correction & info dection; cheap fix & info localization -> provide insight
\begin{abstract}
Speech language models are increasingly evaluated on paralinguistic tasks by the accuracy of prompted answers, but answer accuracy combines failures at different stages of the audio-to-answer computation. We introduce a generation-aligned diagnostic ladder that compares the emitted answer, the option logits, an affine readout of those logits, and a linear readout of the hidden state at the same answer token. Successive differences separate endpoint, decision-rule, and readout-coverage gaps. Across five systems and two emotion corpora, state decoding exceeds generation by 27.8 accuracy points on average, and both the decision-rule and readout-coverage gaps are positive in all ten conditions. A label-free logit correction improves generated accuracy in every condition, showing that part of the decision-rule gap is actionable. In rank-matched comparisons, emotion information outside the native readout generalizes to held-out speakers and survives controls for measured acoustic descriptors, but replacing the selected readout-external directions usually has little effect on emitted answers. These results distinguish information availability from behavioral use and localize performance losses across the decision rule and the state-to-answer readout.
\end{abstract}


\section{Introduction}
% Background. do we need to report the audio-path and layer-wise LLM layer peformenace? No it will complicate our main claim
\IEEEPARstart{S}{peech} language models couple audio interfaces with generative language backbones, allowing a single system to process speech and respond through natural-language generation \cite{qwen25omni,qwen2audio,audioflamingo3,kimiaudio,phi4mm}. Their evaluation has expanded beyond transcription to emotion, prosody, and other paralinguistic judgments, typically through prompted multiple-choice or free-form responses \cite{dynamicsuperb,airbench,sdeval}. Generation accuracy provides a convenient summary of behavioral performance, but it conflates at least three distinct failure locations: the relevant evidence may never reach the language-model component; it may be retained in the state without being exposed through the answer readout; or it may reach the answer logits yet be misread by a mismatched decision rule. These possibilities demand different interventions, yet behavioral accuracy alone cannot distinguish among them.

% Our main ob
We ground this localization problem in four-class speech emotion recognition, whose acoustic correlates are well characterized \cite{scherer2003vocal,juslin2003communication} and whose answer can be elicited as a single token. Across ten system--corpus conditions, generated accuracy averages \(0.474\), whereas linear decoding from the hidden state at the same answer position averages \(0.752\); this deficit of \(0.278\) in absolute accuracy persists under speaker-disjoint evaluation and appears in every condition. It is most pronounced in relative terms for Phi-4-MM on CREMA-D \cite{cao2014cremad}, where generation reaches \(0.279\) against a full-state probe at \(0.722\). These results reveal a substantial state-to-answer loss even when emotion remains linearly decodable at the answer position.

% Two gaps -> tow fixes
This discrepancy points to two distinct failure locations within the language model, which we introduce in the order encountered when tracing back from the emitted answer. First, closest to the answer, emotion evidence may be present in the option logits while the default decision rule uses those scores inefficiently. We call this \emph{decision-rule misalignment}. Second, deeper in the state, additional evidence may remain accessible at the answer position without being expressed through the option-specific coordinates that determine the generated label. We call this a \emph{readout-coverage gap}. The practical consequence is that decision-rule misalignment is amenable to logit correction, whereas a coverage gap cannot be recovered by reweighting the same option logits and requires moving beyond the native readout. Phi-4-MM on CREMA-D makes this concrete. The best affine rule over its option logits reaches an accuracy of only \(0.354\), far below the \(0.722\) supported by the full state, so most of its deficit cannot be repaired at the option logits.

% ladder
To localize these failures, we construct a generation-aligned diagnostic ladder that compares actual generation, the default option-logit decision, an optimized affine decision over the same logits, and regularized linear decoding from the full state. Anchoring all four levels to one verified answer token ensures that behavior, logits, and states measure the same event. Their successive differences split the distance between generated accuracy and full-state decodability into three terms that sum exactly: endpoint validity (agreement between the emitted answer and the option favored by the model's own logits), the decision-rule gap, and the readout-coverage gap. Because the full-state reader has many more input dimensions than the contrast reader, we treat the readout-coverage gap as a performance gap; claims about readout-external information rest on rank-matched comparisons. We then connect diagnosis to behavior. A label-free logit correction tests the decision-rule gap during generation, while held-out decoding and minimal-pair subspace interventions distinguish the availability of readout-external information from its causal use. Acoustic controls characterize how much of that information is explained by measured surface cues.

% contribution
These analyses yield four contributions. First, we introduce a generation-aligned framework that exactly decomposes the distance between emitted behavior and full-state decodability into endpoint validity, a decision-rule gap, and a readout-coverage gap. Second, we show that the decision-rule diagnosis is behaviorally actionable. A standard label-free logit correction \cite{zhao2021calibrate,zhou2024batch} improves the emitted answer across all ten conditions with negligible format cost (Section~\ref{sec:offset}). Third, in rank-matched comparisons, we identify emotion information that remains linearly accessible outside the native readout and generalizes to held-out speakers. Fourth, matched minimal-pair interventions show that the selected readout-external directions have limited influence on the emitted answer, separating information availability from causal use.

\section{Related Work}

\paragraph{Prosody-sensitive evaluation of speech language models}
Speech language models route speech through an encoder and projector into a text LLM \cite{zhang2023speechgpt,qwen2audio,salmonn,audioflamingo,qwen25omni}. Benchmarks such as Dynamic-SUPERB, AIR-Bench, and SD-Eval include emotion and paralinguistic tasks \cite{dynamicsuperb,airbench,sdeval}, and controlled studies show that models often rely more on lexical than acoustic cues \cite{chen2026listen}. These works establish behavioral gaps but do not localize whether a cue is lost, attenuated, or retained but underused. We study speech without engineered text--audio conflict and localize these downstream failures within the audio-to-answer computation.

\paragraph{Label-free correction of class-dependent readout offsets}
Prompted multiple-choice answers exhibit systematic position and token biases \cite{zheng2024mcqbias}. Prior label-free methods estimate such preferences from content-free inputs \cite{zhao2021calibrate}, answer-side scoring statistics \cite{kumar2022answerlevel}, or the mean predicted distribution over an unlabeled batch \cite{zhou2024batch}. We adopt the latter estimator but use it as a behavioral intervention rather than offline rescoring, applying the offset at the first answer step, resuming full-vocabulary generation, and scoring the emitted string. This measures both accuracy gain and format stability, making the correction a behavioral test of the decision-rule gap.

\paragraph{Representation probing in speech models}
Beyond the option scores, linear probing shows that emotion and prosody are recoverable from frozen speech encoders \cite{hubert,wavlm,whisper,pasad2021layerwise,deseyssel2022probing,wagner2023dawn,emotion2vec}. Layer-wise probing inside speech language models further shows that such attributes remain recoverable deep in the language stack \cite{yang2025audiolens}. However, decodability does not imply use \cite{belinkov2022probing}; we therefore treat probing as an availability diagnostic rather than evidence that the complete system recruits the cue.

\paragraph{Subspace interventions and the selection trap}
The logit lens and hidden-state interventions provide tools for localizing computation \cite{nostalgebraist2020lens,belrose2023eliciting,vig2020causal,geiger2021causal,meng2022locating}. Supervised subspace selection, however, can confound decodability with mechanism \cite{makelov2023subspace}. We address this problem by fixing the native readout space from the output head and treating held-out decoding and matched replacement as separate measurements of availability and use. In text-only models, a related knowledge--prediction gap has been reported on multiple-choice questions \cite{park2025kappa}; our framework additionally separates readout coverage from the decision rule over option scores and anchors both to the generated answer. Concurrent work retrieves sparse audio concepts \cite{chowdhury2026ard} and studies text--audio conflict \cite{cho2026conflict}; our focus is the availability and causal use of readout-external information in ordinary prompted speech.

\section{Diagnosing the State-to-Answer Interface}
\label{sec:alignment}

\begin{figure*}[t]
\centering
\includegraphics[width=\linewidth]{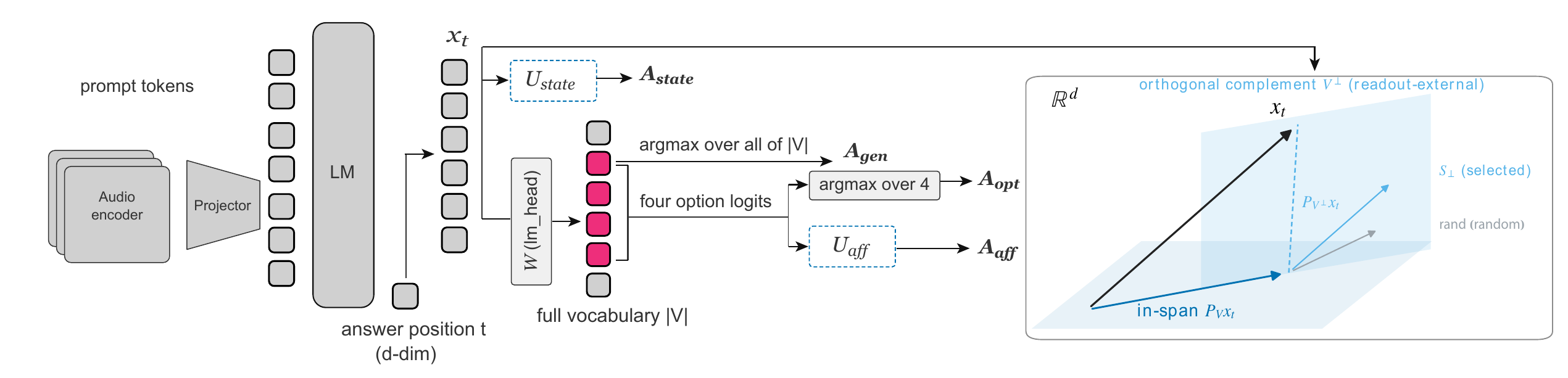}
\caption{\textbf{Overview of the generation-aligned diagnostic ladder.}
Left: audio and prompt tokens are processed by the encoder, projector, and language model, and the output head \(W\) maps the answer-position state \(x_t\) to full-vocabulary logits. Four readouts score the same answer event: greedy generation over the full vocabulary (\(A_{\mathrm{gen}}\)), the post-hoc argmax over the four option logits (\(A_{\mathrm{opt}}\)), a learned affine reader on the option-logit contrasts (\(A_{\mathrm{aff}}\)), and a learned affine reader on the full state (\(A_{\mathrm{state}}\)); dashed boxes mark the learned readers, and successive differences among the four accuracies give the gaps in Eq.~\eqref{eq:ladder}. Right: \(x_t\) decomposes into its component in the answer-readout span (\(P_{V}x_t\)) and the readout-external complement \(V^{\perp}\), within which a supervised-selected subspace \(S_{\mathrm{decoding}}\) is compared against same-rank random subspaces (Sections~\ref{sec:readout-geometry} and~\ref{sec:causal-intervention}).}
\label{fig:schematic}
\end{figure*}

Figure~\ref{fig:schematic} gives an overview of this section. It develops a diagnostic ladder that aligns the emitted token, the option scores, and the underlying hidden state at the same first-token event, and a decomposition of that state around the answer readout. Both concern the last mile; they do not by themselves localize losses earlier in the audio pathway.

\subsection{Three Views for One Answer}
\label{sec:endpoint}

For each audio--prompt pair, let \(x_t\in\mathbb{R}^{d}\) be the model-native, post-normalization hidden state used to predict the first answer token, \(\mathcal{V}\) the vocabulary, and \(W\in\mathbb{R}^{|\mathcal{V}|\times d}\) the language-model output head. For option-token identifiers \(v_1,\ldots,v_4\), define

\begin{equation}
\begin{aligned}
z_t &= Wx_t,\\
z_{t,\mathrm{opt}} &= \big(z_{t,v_1},\ldots,z_{t,v_4}\big),\\
y_t &= \operatorname*{arg\,max}_{v\in\mathcal{V}} z_{t,v},
\end{aligned}
\label{eq:three-views}
\end{equation}

\noindent at the same inference step. Thus, \(y_t\), \(z_{t,\mathrm{opt}}\), and \(x_t\) provide three views of one answer event: emitted behavior, native option preference, and the state available to the readout.

\subsection{A Four-Level Diagnostic Ladder}
\label{sec:ladder}

We operationalize these views as four levels of emotion-classification performance (Fig.~\ref{fig:schematic}, left), all evaluated on the same rows and with the same scoring rule; learned readers are assessed on the same held-out splits:

\begin{itemize}
\item \(A_{\mathrm{gen}}\): accuracy of the model's actual generated answer, with any response outside the required format scored as incorrect;
\item \(A_{\mathrm{opt}}\): accuracy of the post-hoc argmax restricted to the four option logits;
\item \(A_{\mathrm{aff}}\): accuracy of a learned affine reader applied to three reference-relative option-logit contrasts;
\item \(A_{\mathrm{state}}\): accuracy of a regularized affine reader applied to the full answer-position state.
\end{itemize}

An off-option emission or a mismatch in native tie-breaking can make \(A_{\mathrm{gen}}\) differ from \(A_{\mathrm{opt}}\). For the two learned readers, let \(c_t=(z_{t,v_1}-z_{t,v_4},\,z_{t,v_2}-z_{t,v_4},\,z_{t,v_3}-z_{t,v_4})^\top\), and set \(h_{\mathrm{aff}}=c_t\) and \(h_{\mathrm{state}}=x_t\). Both predictions take the form

\begin{equation}
\hat y_r
=\operatorname*{arg\,max}_{k}
  (U_r h_r+b_r)_k,
\qquad
r\in\{\mathrm{aff},\mathrm{state}\}.
\label{eq:learned-readers}
\end{equation}
\(A_{\mathrm{aff}}\) and \(A_{\mathrm{state}}\) are the corresponding held-out accuracies. Both readers are affine and include a bias; \(U\) distinguishes their learned weights from the fixed output head \(W\), and the subscript identifies only whether the input is \(c_t\) or \(x_t\).

Sharing a function class makes the comparison interpretable. The contrast reader can relearn combinations and offsets of the existing option contrasts but cannot access information outside them, whereas the state reader applies the same rule type to the complete state. Their successive differences telescope:

\begin{equation}
A_{\mathrm{state}}-A_{\mathrm{gen}} =
\underbrace{A_{\mathrm{opt}}-A_{\mathrm{gen}}}_{\Delta_{\mathrm{endpoint}}}
+
\underbrace{A_{\mathrm{aff}}-A_{\mathrm{opt}}}_{\Delta_{\mathrm{decision}}}
+
\underbrace{A_{\mathrm{state}}-A_{\mathrm{aff}}}_{\Delta_{\mathrm{coverage}}}.
\label{eq:ladder}
\end{equation}

\(\Delta_{\mathrm{endpoint}}\) is the accuracy difference between emitted generation and the option-only decision. It primarily captures emission and formatting behavior, including off-option responses and native tie-breaking mismatches, so we report it without a separate mechanistic analysis. A near-zero \(\Delta_{\mathrm{endpoint}}\) certifies that the option-restricted view is a faithful anchor for the two explanatory gaps. \(\Delta_{\mathrm{decision}}\), the \emph{decision-rule gap}, measures the gain available from a better rule over the existing contrasts. \(\Delta_{\mathrm{coverage}}\), the \emph{readout-coverage gap}, measures the additional performance supported by the full state. Here, \(A_{\mathrm{state}}\) is a decodability reference rather than attainable model performance. The identity is exact by telescoping, and finite-sample gap estimates are reported with uncertainty. We analyze the two explanatory gaps in ladder order, beginning with the decision-rule gap.

\subsection{Logit Correction of the Decision-Rule Gap}
\label{sec:output-repair}

A positive \(\Delta_{\mathrm{decision}}\) means that an affine rule improves on the native option-only choice using the same contrasts. One possible source is stable option bias \cite{zheng2024mcqbias}. We test it with the unlabeled-batch estimator, part of a broader family of label-free corrections \cite{zhou2024batch,zhao2021calibrate,kumar2022answerlevel}, and score its effect on subsequent generation.

For each utterance \(x\) under a fixed prompt variant, let \(q_k(x)\) be the model's probability for option \(k\), normalized over the four prompted options. Averaging this quantity over a condition- and prompt-matched unlabeled set \(\mathcal{C}\) estimates how strongly the model favors each option overall. We define

\begin{equation}
\hat p_k=\frac{1}{|\mathcal{C}|}\sum_{x_i\in\mathcal{C}}q_k(x_i),
\qquad
b_k=-\log\hat p_k+\frac{1}{4}\sum_{j=1}^{4}\log\hat p_j,
\label{eq:offset}
\end{equation}

\noindent Here, \(\hat p_k\) is the estimated average preference and \(b_k\) reverses and centers it. A frequently favored option receives a smaller offset, whereas an underpreferred option receives a larger one. If all four options are favored equally, every offset is zero, so neither their relative ordering nor their competition with off-option tokens changes. Because a marginal option preference can also reflect the target class distribution, interpreting it as bias requires a target-prior assumption; here the intended target prior is uniform, and the retained four-class subsets are nearly balanced.

We add \(b_k\) to the corresponding option-token logit only at the first answer step and then continue ordinary full-vocabulary generation. Let \(A_{\mathrm{offset}}\) be the resulting accuracy under the same parser used for \(A_{\mathrm{gen}}\). The gain \(A_{\mathrm{offset}}-A_{\mathrm{gen}}\) therefore measures whether this simple correction improves the answers the model actually emits. Because the gain is scored at the generation endpoint, where the offsets also shift the options' competition with off-option tokens, it tests the bias account of \(\Delta_{\mathrm{decision}}\) rather than estimating that gap directly. Ground-truth labels are used only afterward to score the gain; they do not enter the correction. In our experiments, \(\mathcal{C}\) contains the target-batch inputs themselves, making the procedure label-free but transductive.

\subsection{Localizing the Readout-Coverage Gap}
\label{sec:readout-geometry}

A positive \(\Delta_{\mathrm{coverage}}\) shows that the full hidden state supports better linear emotion decoding than the three option-logit contrasts. It does not show where the additional decodable information lies, partly because the full-state reader receives many more input dimensions. This subsection therefore asks whether emotion information is concentrated in the model's native answer readout or also remains available outside it. We use rank-matched comparisons here and report a matched-budget refit in Supplementary Section~\ref{app:robust}.

We first identify the hidden-state directions that can directly change the relative option logits. Using the fourth option as reference, define
\begin{equation}
C =
\begin{bmatrix}
W_{v_1,:}-W_{v_4,:}\\
W_{v_2,:}-W_{v_4,:}\\
W_{v_3,:}-W_{v_4,:}
\end{bmatrix}
\in\mathbb{R}^{3\times d},
\qquad
V_3=\operatorname{row}(C).
\label{eq:contrast}
\end{equation}

\noindent We call \(V_3\) the prompt-specific \emph{answer-readout space} (Fig.~\ref{fig:schematic}, right). It has three dimensions because four option scores have three independent relative contrasts; the rows of \(C\) are linearly independent in the analyzed systems. The output head \(W\) is fixed; the prompt enters only by determining which four option-token rows of \(W\) define \(C\). The contrast reader observes \(c_t=Cx_t\), so it can access only the component of the state in \(V_3\). Let \(P_{V_3}\) and \(P_{V_3^\perp}\) denote the orthogonal projections onto \(V_3\) and its complement. Then
\begin{equation}
x_t=P_{V_3}x_t+P_{V_3^\perp}x_t,
\qquad
CP_{V_3^\perp}x_t=0.
\label{eq:readout-decomposition}
\end{equation}

\noindent Thus, the component in \(V_3\) can directly change the relative logits of the four answer options. Information in \(V_3^\perp\) may still be present in the hidden state, but it cannot directly change these relative logits at the final answer position. Throughout, \emph{readout-external} is used in this geometric sense, meaning outside the span of the option-token rows of the output head, not unrelated to the task.

This decomposition is exact at the final answer position. To study how the same information is organized before the final readout, we move to an intermediate layer \(L^*\). We select \(L^*\) on the training split as the layer with the highest logit-lens accuracy, obtained by applying the model's final normalization and output head to the answer-position state (Section~\ref{sec:setup-subspaces}). At \(L^*\), \(V_3\) is therefore a reference aligned with the final readout, not an exact decomposition of the final logits.

For the external subspaces, we remove the full option-row span \(V_4=\operatorname{row}(W_{v_1:v_4})\). Because \(V_3\subset V_4\), their complements satisfy \(V_4^\perp\subset V_3^\perp\). A direction selected in \(V_4^\perp\) is therefore also external to the relative option readout represented by \(V_3\). The decoding comparisons below use rank-three spaces matched to \(V_3\); the interventions in Section~\ref{sec:causal-intervention} instead use rank-four spaces matched to \(V_4\), which also carries the absolute option logits that matter during unconstrained generation.

We conduct two rank-three comparisons at \(L^*\). First, we compare held-out emotion decoding from \(V_3\) with decoding from random three-dimensional subspaces \(\mathrm{rand}_3\subset V_4^\perp\subset V_3^\perp\). This tests whether the native answer-readout directions carry more emotion information than a random readout-external slice of the same dimension. Neither space is selected using emotion labels, making this the appropriate comparison for evaluating the relative informativeness of the native readout.

Second, we ask whether a generalizable emotion signal can be found outside the native readout. For each prompt, we project the training states onto \(V_4^\perp\subset V_3^\perp\), fit a supervised multinomial logistic model, and define \(S_{\mathrm{decoding}}\) from the three leading right-singular directions of its coefficient matrix. We then fit a decoder in \(S_{\mathrm{decoding}}\) and evaluate it on held-out speakers. An advantage over \(\mathrm{rand}_3\) shows that selected readout-external directions contain emotion information that generalizes beyond the training speakers. Because \(S_{\mathrm{decoding}}\) is selected using emotion labels whereas \(V_3\) is not, their accuracies do not provide a direct ranking of the native and external spaces. Construction details are given in Section~\ref{sec:setup-subspaces} and the supplementary material.

\subsection{Causal Interventions on Readout-External Information}
\label{sec:causal-intervention}

Held-out decoding shows what information is available outside the readout, but not whether that information affects the model's answer. We therefore replace selected components of the answer-position state at \(L^*\) and continue generation. The decoding analysis uses \(V_3\), which represents the three relative contrasts among four options. The intervention instead uses the full option-row span \(V_4\) defined above, because unconstrained generation also depends on the absolute option logits and their competition with off-option tokens.

We compare three rank-four spaces. \(V_4\) is the model's native option-readout space. \(S_{\mathrm{intervention}}\subset V_4^\perp\) contains the four leading supervised readout-external directions. It extends the rank-three decoding space, so \(S_{\mathrm{decoding}}\subset S_{\mathrm{intervention}}\). Finally, \(\mathrm{rand}_4\subset V_4^\perp\) is a same-rank random control. Replacing \(V_4\) tests whether the answer responds to information directly aligned with the native readout. Replacing \(S_{\mathrm{intervention}}\) tests whether selected information outside that readout can influence the answer, while \(\mathrm{rand}_4\) controls for a generic state perturbation.

For each held-out minimal pair, the receiver and donor share the same speaker and transcript but express different emotions. Let \(h_r\) and \(h_d\) be their answer-position states at \(L^*\) under the same prompt. For \(S\in\{V_4,S_{\mathrm{intervention}},\mathrm{rand}_4\}\), we construct
\begin{equation}
\widetilde h_r^{(S)}
=h_r+P_S(h_d-h_r).
\label{eq:subspace-replacement}
\end{equation}
This operation replaces only the receiver's component in \(S\) with the donor's component. We then continue generation from the edited state and measure two outcomes. The answer-change rate \(R_{\mathrm{chg}}(S)\) records any change from the receiver's original answer. The donor-following rate \(R_{\mathrm{don}}(S)\) counts only the cases in which the answer changes to the donor's emotion category, and therefore measures content-specific transfer. Each reported effect is the paired difference from the \(\mathrm{rand}_4\) arm. A full-state replacement checks that downstream generation can respond to a state change at \(L^*\). For the depth analysis, we repeat the same intervention at several layers. Implementation and statistical details are given in Section~\ref{sec:setup-patching}.

\subsection{Controls for Surface Acoustic Confounds}
\label{sec:acoustic-content}

The rank-three analysis tests whether \(S_{\mathrm{decoding}}\) supports held-out emotion decoding outside the native readout. One possible explanation is that this performance is driven mainly by simple surface acoustic cues that covary with the emotion labels. Such cue--label relationships can occur in acted-emotion corpora; for example, overall recording level can itself support decoding \cite{scherer2003vocal,juslin2003communication}. We test this explanation with three controls. First, we decode emotion from clip-level acoustic-prosodic descriptors alone; the resulting accuracy \(A_{\mathrm{desc}}\) measures their predictive strength. Second, we regress the \(S_{\mathrm{decoding}}\) coordinates on those descriptors and decode from the residuals; the accuracy drop measures how much decoding depends on the measured cues. Third, we equalize the loudness of every clip, re-extract the states, and repeat the decoding to test dependence on absolute level. An extended descriptor panel and a nonlinear removal variant provide stronger versions of the same control; panel composition and protocols are given in Section~\ref{sec:setup-acoustics}.

\section{Experimental Setup}
\label{sec:setup}

\subsection{Models, Corpora, and Evaluation}
\label{sec:setup-models}
\label{sec:setup-generation}

We evaluate Qwen2.5-Omni-7B, Qwen2-Audio-7B, Audio-Flamingo-3, Kimi-Audio-7B, and Phi-4-MM \cite{qwen25omni,qwen2audio,audioflamingo3,kimiaudio,phi4mm}. The task is four-way classification of happy, sad, angry, and neutral. CREMA-D \cite{cao2014cremad} contributes 4,900 clips from 91 speakers, and VESUS \cite{sager2019vesus} contributes 10,073 clips from 10 speakers, giving 10 model--corpus conditions.

Four Latin-square prompt variants rotate the emotions through the four option positions. Under each model's tokenizer, every selected option verbalizer is a single native vocabulary token. Generation is greedy over the full vocabulary without masking non-option tokens. A strict prefix parser maps valid answer surfaces to emotion labels; refusals, ambiguous answers, and off-format responses are incorrect. All four ladder levels use the same option rows, with the generated answer, option logits, and hidden state recorded at the same first-answer-token event.

\(A_{\mathrm{gen}}\) and \(A_{\mathrm{opt}}\) require no fitting; \(A_{\mathrm{aff}}\) and \(A_{\mathrm{state}}\) use five speaker-disjoint outer folds. Standardization, penalty selection, and reader fitting are confined to the training speakers in each fold. The observational unit is the clip. Speaker-clustered resampling keeps all clips and prompt variants from one speaker together. We report paired, per-condition 95\% intervals without family-wise adjustment.

\subsection{Subspace Decoding and Causal Replacement}
\label{sec:setup-subspaces}
\label{sec:setup-patching}

We split speakers into fixed training and held-out sets. Training speakers select \(L^*\) by logit-lens accuracy, construct \(S_{\mathrm{decoding}}\), and fit the decoders; held-out speakers are reserved for evaluation. The random-space results average \(\mathrm{rand}_3\) over 20 independent draws and report the spread across draws.

The matched replacements use the speaker split and subspaces defined above. Within each corpus, every system and intervention arm uses the same receiver--donor pairs, and the random arm uses one fixed \(\mathrm{rand}_4\). Effects are paired differences from the random arm, with uncertainty clustered by speaker. We also report the one-sided 95\% upper bound for each readout-external effect as a share of the corresponding readout-aligned effect. A full-state replacement provides a perturbability control at \(L^*\).

\subsection{Controls for Surface Acoustic Confounds}
\label{sec:setup-acoustics}

Surface acoustic cues can covary with emotion labels, so we test whether they explain the held-out decodability of \(S_{\mathrm{decoding}}\). We use a ten-descriptor base panel and a twenty-descriptor extended panel. \(A_{\mathrm{desc}}\) fits the same decoder family to the descriptors alone using the same speaker split. We then regress the \(S_{\mathrm{decoding}}\) coordinates on each descriptor panel and decode from the residuals, with all statistics estimated on training speakers only. The nonlinear variant replaces linear regression with gradient-boosted trees. The input-side control RMS-equalizes each clip, re-extracts the answer-position states under the same prompts and \(L^*\), and repeats the subspace analysis.

The Supplementary Material provides the remaining experimental configuration details, including model checkpoints and prompt templates, tokenizer and endpoint audits, reader fitting and speaker splits, subspace construction and random-space sampling, receiver--donor pairing, and acoustic descriptor definitions and control protocols.

\section{Results}
\label{sec:studies}

We organize the results around the generation-aligned performance ladder. We first report its endpoint, decision-rule, and readout-coverage gaps, then test label-free logit correction, the availability of readout-external information, its causal use, and finally controls for measured surface acoustic cues.

\subsection{Both Gaps Are Systematic, but Their Relative Importance Varies}
\label{sec:two-gaps}

\begin{figure}[t]
\centering
\includegraphics[width=\columnwidth]{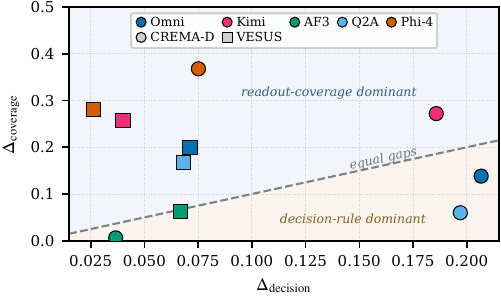}
\caption{\textbf{Which gap dominates differs by condition.} Each point is one model--corpus condition, with color denoting the system and shape the corpus. The dashed line marks equal gaps and the shading separates the two regimes: points in the blue region above the line lose more at the readout (coverage-dominant), and points in the tan region below it lose more at the decision rule (decision-dominant). Exact values and intervals are in Table~\ref{tab:ladder}.}
\label{fig:gapprofile}
\end{figure}

Table~\ref{tab:ladder} reports all three performance-ladder terms across the ten conditions, whose mean \(A_{\mathrm{gen}}\) is \(0.474\). \(\Delta_{\mathrm{endpoint}}\) is effectively zero throughout. The emitted answer achieves the same accuracy as the option favored by the model's logits, so no meaningful performance is lost at this interface. Both explanatory gaps are positive. \(\Delta_{\mathrm{decision}}\) ranges from \(+0.0264\) to \(+0.2067\), showing that the default decision over the option logits falls short of a fitted rule on those same logits. \(\Delta_{\mathrm{coverage}}\) ranges from \(+0.0062\) to \(+0.3679\), showing that the fitted logit rule in turn falls short of a reader of the full answer-position state. The confidence intervals for both gaps exclude zero in every condition.

Figure~\ref{fig:gapprofile} compares the relative sizes of the two explanatory gaps. Qwen2-Audio \(\times\) CREMA-D is decision-dominant, with \(\Delta_{\mathrm{decision}}\) at \(+0.197\) compared with a \(+0.060\) readout-coverage gap. Phi-4-MM \(\times\) CREMA-D is coverage-dominant; its \(+0.368\) readout-coverage gap is the largest in the study, whereas \(\Delta_{\mathrm{decision}}\) is \(+0.075\). Qwen2.5-Omni \(\times\) CREMA-D has substantial losses at both transitions, including the largest \(\Delta_{\mathrm{decision}}\) (\(+0.207\)) and a \(+0.138\) readout-coverage gap. The remaining conditions lie between these patterns, with Kimi-Audio showing the widest overall separation between generated behavior and state decodability.

% Main performance ladder, 10 model-corpus conditions.
% Numbers: FRAMING.md section 14.1 ledger (= EXP_BREADTH_RESULTS.md section 4.1);
% same run as the auto-generated tab_breadth_ladder.tex, restated in the
% manuscript's own notation (A_gen / A_opt / A_aff / A_state instead of R0/R1/R3/R5).
% The telescoping residual is exactly 0 in every row.
\begin{table*}[t]
\centering
\scriptsize
\setlength{\tabcolsep}{5pt}
\caption{The distance between generated behavior and linear decodability decomposes into an endpoint-validity term and two explanatory gaps. Rows are ordered by \(A_{\mathrm{gen}}\). \(\Delta_{\mathrm{decision}}=A_{\mathrm{aff}}-A_{\mathrm{opt}}\) and \(\Delta_{\mathrm{coverage}}=A_{\mathrm{state}}-A_{\mathrm{aff}}\).}
\label{tab:ladder}
\begin{tabular}{llrrrrll}
\toprule
System & Corpus & \(A_{\mathrm{gen}}\) & \(A_{\mathrm{opt}}\) & \(A_{\mathrm{aff}}\) & \(A_{\mathrm{state}}\) & \(\Delta_{\mathrm{decision}}\) [95\% CI] & \(\Delta_{\mathrm{coverage}}\) [95\% CI] \\
\midrule
Audio-Flamingo-3 & CREMA-D & 0.9060 & 0.9060 & 0.9426 & 0.9487 & $+0.0366$\,[$+0.0289,+0.0446$] & $+0.0062$\,[$+0.0022,+0.0104$] \\
Qwen2-Audio      & CREMA-D & 0.7025 & 0.7025 & 0.8995 & 0.9594 & $+0.1970$\,[$+0.1909,+0.2030$] & $+0.0599$\,[$+0.0540,+0.0658$] \\
Audio-Flamingo-3 & VESUS   & 0.5667 & 0.5667 & 0.6335 & 0.6955 & $+0.0668$\,[$+0.0244,+0.1065$] & $+0.0620$\,[$+0.0122,+0.1037$] \\
Qwen2.5-Omni     & CREMA-D & 0.5265 & 0.5265 & 0.7332 & 0.8714 & $+0.2067$\,[$+0.1933,+0.2190$] & $+0.1382$\,[$+0.1275,+0.1492$] \\
Kimi-Audio       & CREMA-D & 0.3994 & 0.3994 & 0.5852 & 0.8572 & $+0.1858$\,[$+0.1741,+0.1977$] & $+0.2720$\,[$+0.2577,+0.2857$] \\
Kimi-Audio       & VESUS   & 0.3704 & 0.3704 & 0.4104 & 0.6684 & $+0.0400$\,[$+0.0263,+0.0530$] & $+0.2581$\,[$+0.2277,+0.2921$] \\
Qwen2-Audio      & VESUS   & 0.3677 & 0.3677 & 0.4358 & 0.6029 & $+0.0681$\,[$+0.0501,+0.0888$] & $+0.1671$\,[$+0.1451,+0.1909$] \\
Qwen2.5-Omni     & VESUS   & 0.3590 & 0.3590 & 0.4302 & 0.6290 & $+0.0712$\,[$+0.0508,+0.0917$] & $+0.1988$\,[$+0.1635,+0.2345$] \\
Phi-4-MM         & CREMA-D & 0.2793 & 0.2793 & 0.3544 & 0.7223 & $+0.0751$\,[$+0.0685,+0.0820$] & $+0.3679$\,[$+0.3508,+0.3838$] \\
Phi-4-MM         & VESUS   & 0.2594 & 0.2595 & 0.2859 & 0.5660 & $+0.0264$\,[$+0.0130,+0.0401$] & $+0.2801$\,[$+0.2419,+0.3250$] \\
\bottomrule
\end{tabular}
\end{table*}

Robustness analyses reproduce the gaps with nonlinear logit-side decoding, matched regularization budgets, and delete-one-speaker resampling; complete results are in Supplementary Section~\ref{app:robust}.

The ladder reveals no single universal failure profile. Some conditions have a larger decision-rule gap, others have a larger readout-coverage gap, and several show substantial gaps at both transitions.

\subsection{Label-Free Logit Correction Recovers Part of the Decision-Rule Gap}
\label{sec:offset}

\begin{figure}[t]
\centering
\includegraphics[width=\columnwidth]{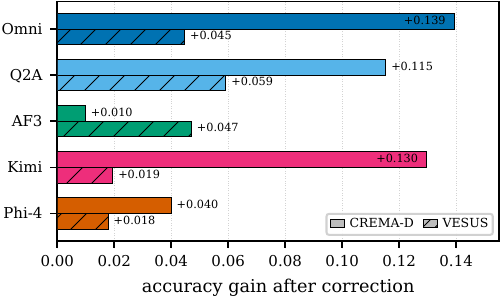}
\caption{\textbf{A label-free logit correction, measured on the emitted answer.} Gains are grouped by system, with a solid CREMA-D bar and a hatched VESUS bar. Bars show only the generated-accuracy gain over the uncorrected baseline; baseline and corrected accuracies, together with transmission and parseability diagnostics, are in Supplementary Table~\ref{tab:app-offset}.}
\label{fig:offset}
\end{figure}

Figure~\ref{fig:offset} shows that label-free logit correction improves the emitted answer in all ten conditions, with accuracy gains from \(+0.0098\) to \(+0.1392\) (per-condition values in Supplementary Table~\ref{tab:app-offset}). These generation-time gains recover part of the decision-rule gap without ground-truth labels. The largest gain is \(+0.1392\) on Qwen2.5-Omni \(\times\) CREMA-D. The correction also helps across performance regimes. Phi-4-MM gains \(+0.0399\) and \(+0.0178\) with its coverage-dominant profile, while Audio-Flamingo-3 gains \(+0.0098\) on CREMA-D from a 0.9060 baseline. Across conditions, the offset realizes 0.27 to 0.87 of the supervised decision-rule gap. The label-free correction is therefore effective across all tested model--corpus conditions, although the size of the gain varies. Because the offsets are estimated from the unlabeled evaluation batch itself, the correction is transductive; applying it to a single isolated example would require other target-domain data (Section~\ref{sec:limitations}).

Because the offsets modify logits during unconstrained generation, they could also make the model produce answers outside the required format. We therefore check whether the corrected option is actually emitted and whether the answer remains parseable. The corrected option is emitted on 95.53\% to 100\% of rows, with parseability unchanged except for a 0.0019 loss on Qwen2.5-Omni \(\times\) VESUS. The remaining difference from the supervised gap indicates that systematic option-prior bias is one contributor rather than its complete explanation.

\subsection{Readout-External Emotion Information Generalizes}
\label{sec:availability}

\begin{figure*}[t]
\centering
\includegraphics[width=0.98\textwidth]{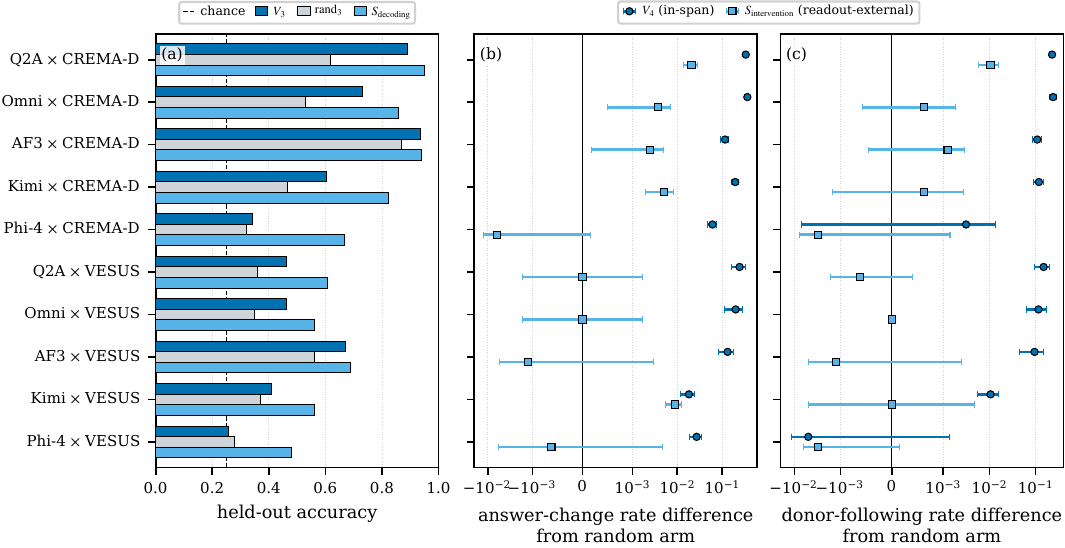}
\caption{\textbf{Availability does not imply effective use.} (a) Held-out decoding accuracy at \(L^*\) for the native readout space \(V_3\), a label-free random subspace \(\mathrm{rand}_3\subset V_4^{\perp}\subset V_3^{\perp}\) averaged over 20 draws, and the supervised space \(S_{\mathrm{decoding}}\) selected in the same complement; the dashed line marks chance. Panels (b) and (c) score the rank-four minimal-pair replacements at \(L^*\), shown as paired differences from the \(\mathrm{rand}_4\) arm for \(V_4\) (circles, dark blue) and \(S_{\mathrm{intervention}}\) (squares, light blue). Panel (b) counts any answer change, whereas panel (c) counts only changes to the donor emotion. Bars are speaker-clustered 95\% intervals, and the horizontal axis is symmetric-logarithmic. All panels share their row order. Exact values for (a), (b), and complete donor-following values for (c) are in Supplementary Tables~\ref{tab:subspace}, \ref{tab:patching}, and~\ref{tab:app-patch-follow}.}
\label{fig:availuse}
\end{figure*}

Having tested the decision-rule gap at the output, we turn to the readout-coverage gap and ask whether generalizable emotion information remains decodable outside the native option readout. Fig.~\ref{fig:availuse}(a) shows the held-out decoding results for the native, random, and selected readout-external subspaces, with exact values in Supplementary Table~\ref{tab:subspace}. At the training-selected \(L^*\), held-out decoding from \(S_{\mathrm{decoding}}\subset V_4^\perp\subset V_3^\perp\) reaches 0.481 to 0.949, and its improvement over \(\mathrm{rand}_3\) has a speaker-clustered interval excluding zero in every condition. The answer-position state therefore retains linearly accessible emotion information outside the native option contrasts. Phi-4-MM illustrates the distinction. \(S_{\mathrm{decoding}}\) reaches 0.668 on CREMA-D and 0.481 on VESUS, whereas \(V_3\) reaches only 0.342 and 0.259.

The selection-matched \(V_3-\mathrm{rand}_3\) contrast measures how informative the native readout is relative to a same-rank random space. It is positive with an interval excluding zero in nine of the ten conditions, ranging from \(+0.037\) to \(+0.272\). The exception is Phi-4-MM \(\times\) VESUS at \(-0.018\) [\(-0.033,+0.001\)], and its CREMA-D condition is only \(+0.022\) [\(+0.004,+0.040\)], so Phi-4-MM's native option contrasts carry little more emotion information than a random subspace of the same rank. The next smallest contrast is Kimi-Audio \(\times\) VESUS at \(+0.037\) [\(+0.026,+0.047\)], which shows that a weakly informative native readout is not confined to one language-model family.

Raw accuracy contrasts compress near ceiling. Audio-Flamingo-3 \(\times\) CREMA-D's \(+0.068\) difference, for example, sits on a 0.869 random-space baseline. Individual \(\mathrm{rand}_3\) draws vary by up to 0.05; per-draw results and the \(S_{\mathrm{decoding}}-\mathrm{rand}_3\) intervals are in Supplementary Section~\ref{app:subspaces}.

\subsection{Minimal-Pair Interventions Reveal Limited Causal Use}
\label{sec:causal-use}

Held-out decodability establishes availability; the matched interventions now measure use. Fig.~\ref{fig:availuse}(b,c) shows both outcomes. Replacing the readout-aligned \(V_4\) component increases answer change over the random arm in all ten conditions, by \(+0.0181\) to \(+0.3631\). It also significantly increases donor following in eight conditions; the two Phi-4-MM conditions are nonsignificant. These results show that the model's answer is causally sensitive to changes in the native readout space.

By contrast, replacing the readout-external \(S_{\mathrm{intervention}}\) component has much less influence on the answer. It produces a significant answer-change effect in five conditions, but no effect exceeds \(+0.0206\). Only one condition shows a significant increase in donor following. These results indicate that the selected readout-external information has only limited influence on the emitted answer.

We perform two additional tests to rule out the possibility that the weak \(S_{\mathrm{intervention}}\) effects arise only from small edits or an unresponsive downstream pathway. Exact values, per-condition \(L^*\), and remaining diagnostics are in Supplementary Table~\ref{tab:patching} and Section~\ref{app:patching}.

\begin{figure}[t]
\centering
\includegraphics[width=0.95\columnwidth]{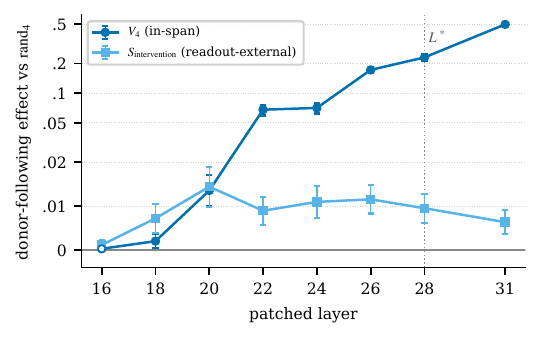}
\caption{\textbf{Depth profile of causal access in Qwen2-Audio \(\times\) CREMA-D.} Donor-following effects of matched rank-four minimal-pair replacement at eight depths, with speaker-clustered 95\% intervals; hollow markers mark intervals containing zero, and the vertical axis is symmetric-logarithmic. Exact values are in Supplementary Section~\ref{app:minlayer}.}
\label{fig:depth-profile}
\end{figure}

Qwen2-Audio \(\times\) CREMA-D is the only condition with a significant readout-external donor-following effect at \(L^*\) (\(+0.0106\)\,[\(+0.0058,+0.0154\)]). We therefore repeat the intervention at eight layers to ask where this causal influence is strongest; Fig.~\ref{fig:depth-profile} shows the resulting donor-following effects. When the replacement is applied at layer 16, the external effect is near zero. It peaks at layer 20, where it matches the readout-aligned effect at the same layer, and becomes weaker when the replacement is applied closer to the output. By contrast, the readout-aligned effect grows toward the output and reaches \(+0.4983\). In this condition, readout-external information has its strongest causal influence in the middle of the model, whereas readout-aligned information becomes increasingly influential near the final readout.

We further track the particular component injected at layer 20 to determine whether it reaches the final answer. Relative to the random control, this intervention shifts the option logits toward the donor emotion by \(+0.190\)\,[\(+0.176,+0.204\)]. However, the shift is usually too small to change which option has the highest score. The mid-stack external pathway therefore reaches the final logits but rarely changes the emitted answer. This depth pattern is established only for Qwen2-Audio \(\times\) CREMA-D (protocols and exact values in Supplementary Sections~\ref{app:minlayer} and~\ref{app:patchtrack}).

These interventions show that readout-external emotion information has limited causal access at the answer.

\subsection{Readout-External Decodability Persists under Controls for Measured Surface Cues}
\label{sec:acoustic-char}

\begin{figure}[t]
\centering
\includegraphics[width=\columnwidth]{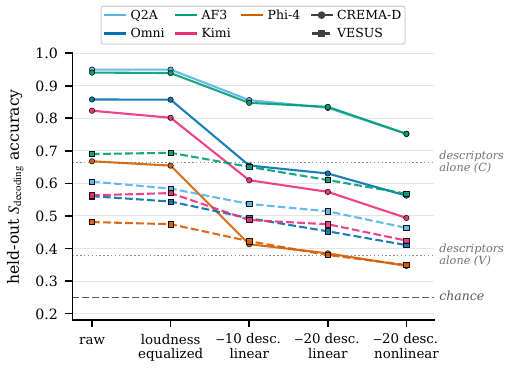}
\caption{\textbf{Readout-external decodability under progressively stronger acoustic controls.} Each line is one model--corpus condition, with color denoting the system, solid circles CREMA-D, and dashed squares VESUS. Shown is held-out \(S_{\mathrm{decoding}}\) accuracy for the raw states, after input loudness equalization, and after residualizing ten or twenty acoustic-prosodic descriptors linearly or with gradient-boosted trees. Dotted lines mark the descriptor-only decoder for each corpus and the dashed gray line marks chance. Exact values and intervals are in Supplementary Tables~\ref{tab:acoustic-char}, \ref{tab:app-sperp-ext}, and~\ref{tab:app-loudnorm-np}.}
\label{fig:acoustic-controls}
\end{figure}

The causal interventions above test whether readout-external information affects the answer. Here we return to the rank-three \(S_{\mathrm{decoding}}\) used in Fig.~\ref{fig:availuse}(a) and ask whether its held-out decodability can be explained by measured surface cues. Recording level alone separates several class pairs in these corpora (Supplementary Table~\ref{tab:app-stimulus}), making shallow clip statistics a plausible source of readout-external decodability. Figure~\ref{fig:acoustic-controls} applies the controls defined in Section~\ref{sec:setup-acoustics}.

Linear residualization of the measured cues reduces held-out \(S_{\mathrm{decoding}}\) accuracy by 0.039 to 0.255. After nonlinear residualization of the extended descriptor panel, accuracy remains at 0.347 to 0.752, above the 0.25 chance level in every condition. Input-side loudness equalization changes accuracy by at most 0.022. Exact values are in Supplementary Tables~\ref{tab:acoustic-char}, \ref{tab:app-sperp-ext}, and~\ref{tab:app-loudnorm-np}. Thus, the measured surface acoustic cues and absolute recording level do not fully explain the held-out decodability of \(S_{\mathrm{decoding}}\).

\section{Discussion}
\label{sec:discussion}

The central result is not simply that a hidden-state decoder outperforms the generated answer. The diagnostic ladder identifies two downstream gaps between information available at the answer position and the answer the model emits. One arises when the model converts its option logits into a choice; the other arises because the option contrasts expose only part of the information available in the hidden state. Both gaps are positive in every evaluated condition, so a system can suffer from both problems at once. They are aggregate properties of a model--corpus condition rather than mutually exclusive explanations of individual errors. More broadly, low generated accuracy need not mean that the relevant emotion evidence never reached the language model.

\subsection{Why the Decision Rule Loses Available Evidence}

Option logits combine evidence from the audio with option-token priors, positional preferences, and prompt-conditioned response habits. A stable preference for one option can therefore shift the default argmax even when the relative scores still contain useful emotion evidence \cite{zheng2024mcqbias}. This interpretation is consistent with the label-free correction. Estimating and removing marginal option preferences improves generation in all ten conditions and recovers 0.27 to 0.87 of the supervised decision-rule gap. The three largest decision-rule gaps also produce the three largest correction gains. Thus, the gap is not only diagnostic; it indicates when a logit-side repair is likely to help.

The correction recovers only part of the gap because a fixed offset captures only the stable component of the mismatch. The remaining difference from the supervised affine reader may reflect class-dependent scaling or example-dependent boundaries that marginal option frequencies cannot estimate. This account predicts that reducing stable option preferences, through prompts or training procedures, should reduce both the decision-rule gap and the benefit of the offset correction.

\subsection{Why Decodable Information Has Limited Influence}

The readout-coverage results reveal a different mismatch. Emotion remains decodable from selected directions outside the native option readout and generalizes to held-out speakers, yet replacing those directions usually has little effect on the answer (no answer-change effect exceeds \(+0.0206\); Section~\ref{sec:causal-use}). A plausible explanation is objective mismatch. Audio front ends and intermediate language-model states may preserve rich prosodic information, while next-token and audio-instruction training reward only the information needed to produce the target text. They do not directly require all emotion-discriminative directions to align with the few output directions that separate the prompted option tokens. A supervised decoder is explicitly trained to find such directions; the native readout is not.

On this account, readout-external information can persist as a usable representation without being part of the model's normal answer pathway. This explains why decodability and behavioral influence diverge, a distinction that probing studies must preserve \cite{belinkov2022probing,xu2020vinformation}. It is especially important here because \(S_{\mathrm{decoding}}\) and \(S_{\mathrm{intervention}}\) are selected with emotion labels. Successful decoding shows that the information is available to a supervised linear reader, not that the model naturally uses the same directions \cite{makelov2023subspace}. The acoustic controls further show that the measured surface cues do not fully explain this decodability, although unmeasured acoustic properties may still contribute.

The depth profile offers a more specific hypothesis about routing. In Qwen2-Audio \(\times\) CREMA-D, readout-external replacement has its strongest donor-directed effect in the middle of the stack and becomes weaker near the output, while the readout-aligned effect grows. This pattern is consistent with progressive consolidation, in which intermediate layers can use emotion information in several directions while later layers increasingly concentrate behaviorally relevant information into token-aligned coordinates. Information that is not transferred into those coordinates may be overwritten or lose access to the answer. Because this pattern is established in one condition, it is best treated as a mechanism to test across models rather than a universal depth profile.

The weak low-rank replacement effects do not imply that every direction outside the readout is functionless. They show that the selected linear component has limited causal access under the tested intervention. Emotion information could also be distributed across more directions or participate through nonlinear interactions that a rank-four replacement does not capture. The strong response to readout-aligned and full-state replacements nevertheless shows that the downstream pathway can respond to state changes at the intervention layer; the main limitation lies in how the selected external information is routed.

\subsection{Implications for Repair and Evaluation}

The two gaps suggest different repairs. A decision-rule gap can be addressed by calibrating the existing option logits, without changing the hidden representation. A readout-coverage gap instead requires changing how hidden-state information reaches the answer, for example through a learned readout adapter, targeted fine-tuning of the output mapping, or auxiliary supervision that aligns prosodic evidence with the option contrasts. If such training reduces \(\Delta_{\mathrm{coverage}}\) and increases the causal effect of readout-external directions, it would support the routing explanation above.

These results also change how generative paralinguistic systems should be evaluated. Generated answers, option logits, and hidden states should be measured at the same answer event; diagnostic readers should be evaluated on identity-disjoint splits; and claims about information use should include dimension-matched causal controls. Reporting these views alongside accuracy separates a failure to represent emotion from a failure to expose or use information that is already present at the answer position.

\section{Limitations}
\label{sec:limitations}

\paragraph{Empirical and statistical coverage}
We evaluate only five models. Although they use several audio front ends, they cover only two language-model families, and four use Qwen-family backbones. Our experiments focus on four-way English emotion classification on CREMA-D and VESUS, using single-token multiple-choice answers. Results may differ for other architectures, spontaneous speech, other languages, other paralinguistic tasks, or free-form answers. VESUS contains only 10 speakers, which limits the precision of speaker-clustered estimates.

\paragraph{Diagnostic scope}
\(A_{\mathrm{aff}}\) reads only three option-logit contrasts, whereas \(A_{\mathrm{state}}\) reads the full \(d\)-dimensional hidden state. Although both use linear classifiers, the state reader has access to many more input dimensions. A larger \(\Delta_{\mathrm{coverage}}\) can therefore arise for two reasons: the hidden state may contain information that the native readout does not expose, or the full-state reader may benefit from its larger input space. We therefore interpret \(\Delta_{\mathrm{coverage}}\) as a performance gap rather than a pure measure of readout geometry.

\paragraph{Acoustic interpretation}
Our acoustic controls account for recording level and twenty measured acoustic descriptors, but they do not cover every property of the speech signal. The remaining decodable information may therefore still include acoustic cues that we did not measure. It should not, by itself, be interpreted as an abstract representation of emotion.

\paragraph{Correction and intervention scope}
The logit correction does not use emotion labels, but it estimates its offsets from the full unlabeled evaluation batch. It therefore requires a batch of examples from the target domain and must be recalibrated for a new domain. It cannot be applied to one isolated example without other target-domain data. The matched replacements are diagnostic tests rather than a trained repair. They ask whether replacing a selected part of the hidden state can change the model's answer; they do not teach the model a new readout or routing mechanism. The one significant readout-external donor-following effect therefore shows only a small amount of causal influence. It does not close the readout-coverage gap.

\section{Conclusion}
\label{sec:conclusion}

Behavioral errors alone do not reveal where task information stops influencing a generated answer. By aligning behavior, option logits, and hidden states at the same first-token event, our diagnostic ladder separates endpoint validity, the decision-rule gap, and the readout-coverage gap. Both explanatory gaps are positive in all ten evaluated conditions. A label-free logit correction recovers part of the decision-rule gap during generation. Meanwhile, emotion information remains decodable outside the native readout under controls for measured acoustic cues, but replacing the selected external component seldom moves the answer toward the donor emotion; this bounds the causal influence of the directions we selected, not of all readout-external information. These results distinguish information availability from behavioral use and motivate different responses: correcting the rule over existing option logits or adapting how hidden-state information is read out and routed. The same three views can be captured at any prompted answer token; in speech, they show where paralinguistic information available at the answer position stops contributing to the emitted response. Reported alongside accuracy, they turn a benchmark score into a diagnosis of the state-to-answer interface.

\bibliographystyle{IEEEtran}
\bibliography{IEEEabrv,custom}

\ifdefined\withpreprintappendix
  \clearpage
  \section*{Supplementary Material}
  \addcontentsline{toc}{section}{Supplementary Material}
  \setcounter{section}{0}
  \setcounter{figure}{0}
  \setcounter{table}{0}
  \setcounter{equation}{0}
  \renewcommand{\thesection}{S\arabic{section}}
  \renewcommand{\thefigure}{S\arabic{figure}}
  \renewcommand{\thetable}{S\arabic{table}}
  \renewcommand{\theequation}{S\arabic{equation}}
  \def\supplyincludedinpreprint{1}
  \ifdefined\supplyincludedinpreprint
  \let\supplyenddocument\relax
\else
\def\supplyenddocument{\end{document}}
\documentclass[journal]{IEEEtran}
\usepackage{amsmath,amsfonts,amssymb}
\usepackage{algorithmic}
\usepackage{algorithm}
\usepackage{array}
\usepackage[caption=false,font=normalsize,labelfont=sf,textfont=sf]{subfig}
\usepackage{textcomp}
\usepackage{stfloats}
\usepackage{url}
\usepackage{verbatim}
\usepackage{graphicx}
\usepackage{cite}
\usepackage{booktabs}
\usepackage{xcolor}
\usepackage{xr}                 % cross-reference the main paper PDF
% Import main.tex's labels (sec:*, main figs/tables) but NOT its \bibcite
% entries: both documents cite the same custom.bib, so importing the main
% paper's bibcites would otherwise raise "multiply-defined" warnings on every
% shared citation key. Neutralize \bibcite only during the import, then restore.
\makeatletter
\let\sv@bibcite\bibcite
\let\bibcite\@gobbletwo
\externaldocument{main}        % imports main.tex's labels (sec:*, main figs/tables)
\let\bibcite\sv@bibcite
\makeatother

% --- Number everything with an "S" prefix so it is unambiguous in the main text ---
\renewcommand{\thesection}{S\arabic{section}}
\renewcommand{\thefigure}{S\arabic{figure}}
\renewcommand{\thetable}{S\arabic{table}}
\renewcommand{\theequation}{S\arabic{equation}}

\begin{document}

% Keep this title in sync with \title in main.tex.
\title{Supplementary Material:\\
Separating Decision-Rule Misalignment from\\ Readout-Coverage Limitations in Speech Language Models}
\author{Linkai~Peng~and~Baorian~Nuchged}
\markboth{Supplementary Material}{}
\maketitle
\fi

\section{Experimental Setup Details}
\label{app:setup}

\subsection{Systems}
\label{app:models}

Table~\ref{tab:app-models} lists the five systems evaluated in the paper, with the checkpoint, architecture, and parameter count of each. For compactness, subsequent supplementary tables abbreviate them as Omni, Q2A, AF3, Kimi, and Phi-4.

We selectively evaluate Qwen2.5-Omni \cite{qwen25omni}, Phi-4-MM \cite{phi4mm}, Audio-Flamingo-3 \cite{audioflamingo3}, Qwen2-Audio \cite{qwen2audio}, and Kimi-Audio \cite{kimiaudio}. These systems are frequently represented in related work and span several widely used speech-language-model architectures. All five expose hidden states along a speech-understanding, text-response pathway and provide the discrete answer endpoint required by our analysis.

\begin{table*}[t]
\scriptsize
\centering
\setlength{\tabcolsep}{3pt}
\caption{The five evaluated systems. Parameter counts are taken from the public model cards and are total architecture sizes, including text/decoder branches where applicable.}
\label{tab:app-models}
\begin{tabular}{lllr}
\toprule
System & HuggingFace checkpoint & Architecture & \#~Params \\
\midrule
Qwen2.5-Omni-7B   & \texttt{Qwen/Qwen2.5-Omni-7B}                & Whisper-style audio tower + 28-layer LLM & 7\,B   \\
Audio-Flamingo-3  & \texttt{nvidia/audio-flamingo-3-hf}          & Whisper-large-v3 + 28-layer Qwen2.5-7B LLM & 8.3\,B \\
Phi-4-MM          & \texttt{microsoft/Phi-4-multimodal-instruct} & Conformer audio tower + 32-layer LLM     & 5.6\,B \\
Qwen2-Audio-7B    & \texttt{Qwen/Qwen2-Audio-7B-Instruct}        & Whisper-large-v3 audio tower + 32-layer Qwen2-7B LLM & 8.4\,B \\
Kimi-Audio-7B     & \texttt{moonshotai/Kimi-Audio-7B-Instruct}   & Whisper-large-v3 encoder + 28-layer Qwen2.5-7B LLM & 9.8\,B \\
\bottomrule
\end{tabular}
\end{table*}

\subsection{Corpus Composition}
\label{app:dataset-distributions}

The clip counts in Table~\ref{tab:dataset-distributions} are filtered subsets of the published corpora, not the full releases.

\begin{itemize}
\setlength{\itemsep}{2pt}
\item \textbf{CREMA-D.} We use the four-emotion subset \(\{\)angry, happy, sad, neutral\(\}\) of the published six-emotion corpus, dropping \emph{disgust} and \emph{fear}, and keep every clip of the retained classes. The source corpus contains fewer neutral clips than non-neutral clips, so the resulting 4,900-clip subset is not class-balanced: its majority class accounts for 25.94\% of clips, against the 25\% four-way chance rate.
\item \textbf{VESUS.} We use the four-emotion subset \(\{\)happy, sad, angry, neutral\(\}\) of the five-emotion corpus, dropping \emph{fearful}. VESUS reads a phonetically balanced, semantically neutral script of more than 250 short phrases, each spoken by 10 actors in every emotion, giving 10,073 clips whose majority class accounts for 25.01\%.
\end{itemize}

Because neither subset is exactly balanced, we report the empirical majority-class rates rather than treating 25\% as an exact baseline.

\begin{table}[t]
\small
\centering
\setlength{\tabcolsep}{4pt}
\caption{Per-class clip counts of the two evaluation corpora.}
\label{tab:dataset-distributions}
\begin{tabular}{lrrrrr}
\toprule
Corpus & happy & sad & angry & neutral & total \\
\midrule
CREMA-D & 1{,}271 & 1{,}271 & 1{,}271 & 1{,}087 & 4{,}900 \\
VESUS   & 2{,}518 & 2{,}517 & 2{,}519 & 2{,}519 & 10{,}073 \\
\bottomrule
\end{tabular}
\end{table}

\subsection{Prompt Protocol}
\label{app:prompts}

\paragraph{System prompts.}
All five systems are queried with their official chat templates, unmodified. Qwen2.5-Omni receives the canonical system message distributed with the model: ``You are Qwen, a virtual human developed by the Qwen Team, Alibaba Group, capable of perceiving auditory and visual inputs, as well as generating text and speech.'' Phi-4-MM's speech-understanding template (\texttt{<|user|>}\allowbreak\texttt{<|audio\_1|>}\allowbreak\texttt{\ldots}\allowbreak\texttt{<|end|>}\allowbreak\texttt{<|assistant|>}) does not include a system turn by design. Audio-Flamingo-3, Qwen2-Audio, and Kimi-Audio are each queried with a single user turn carrying the audio and the text prompt; for these we do not add a system message, following each model's recommended speech-understanding format.

\paragraph{Prompt bank.}
Every clip is presented under the four prompt variants of Table~\ref{tab:prompts-emotion4cls}. They share one instruction template and differ only in the Latin-square assignment of emotions to option letters, which places each emotion at each letter position exactly once across the four variants and so controls for positional bias. The four variants of a clip are repeated measurements of the same audio and are kept together in the same split and the same bootstrap cluster, as detailed in Section~\ref{app:fitting}.

\begin{table}[t]
\scriptsize
\centering
\setlength{\tabcolsep}{3pt}
\renewcommand{\arraystretch}{1.15}
\caption{The four-variant emotion prompt bank. The Latin-square design guarantees that each emotion appears at each letter position exactly once across B1--B4.}
\label{tab:prompts-emotion4cls}
\begin{tabular}{lp{0.7\columnwidth}}
\toprule
ID & Prompt text \\
\midrule
\multicolumn{2}{p{0.92\columnwidth}}{Instruction template: ``Listen to the audio and identify the speaker's emotion. (A) \{A\} (B) \{B\} (C) \{C\} (D) \{D\}. Answer with just the letter A, B, C, or D:''} \\
\addlinespace
B1 & A = happy, B = sad, C = angry, D = neutral \\
B2 & A = sad, B = angry, C = neutral, D = happy \\
B3 & A = angry, B = neutral, C = happy, D = sad \\
B4 & A = neutral, B = happy, C = sad, D = angry \\
\bottomrule
\end{tabular}
\end{table}

\paragraph{Output parsing.}
Generation is greedy over the full vocabulary, and the processor never masks non-option tokens. The parser locates a standalone option marker in leading position, such as \texttt{A}, \texttt{(A)}, or \texttt{A.}, and the corresponding patterns for B, C, and D. Markers are bounded by non-letter characters so that letters inside words such as ``Answer'' are not matched, and the selected option is mapped to the emotion label assigned under the active Latin-square variant. Any generation without a valid leading option marker, including refusals and free-form prose, is counted as incorrect in every main-text analysis; it is never re-parsed into a class by scanning the rest of the response.

\paragraph{Layer-selection readout.}
The criterion that fixes \(L^*\) (Section~\ref{sec:setup-subspaces}) uses a separate readout from the option letters that the analyses score. At each layer, the model's own final normalization and unembedding are applied to the answer-position state, and the score of a class is the log-sum-exp over that class's emotion-word tokens: \(\{\)~happy, ~Happy, ~joyful\(\}\) for happy, \(\{\)~sad, ~Sad, ~upset\(\}\) for sad, \(\{\)~angry, ~Angry\(\}\) for angry, and the corresponding set for neutral. Each listed verbalizer is a single native vocabulary token under the corresponding model tokenizer. The highest-scoring class is the layerwise prediction, and \(L^*\) is the layer with the highest accuracy on the training split. These emotion words are never scored as answers; they enter only the layer-selection criterion.

\subsection{Readers, Splits, and Uncertainty}
\label{app:fitting}

All ladder levels are evaluated on identical rows. The affine and state readers
use five speaker-disjoint outer folds, with feature standardization and every
supervised fitting decision confined to the training side. The affine reader
selects its penalty by nested cross-validation, whereas the main state reader
uses a fixed penalty. Thus the tuned reader is the one subtracted in
\(\Delta_{\mathrm{coverage}}\); Section~\ref{app:robust} reports the refit in
which both readers receive the same nested selection budget.

The audio clip is the observational unit. A resampled speaker carries all of
that speaker's clips and all four prompt variants, preserving their dependence.
Intervals are paired and reported per condition without family-wise
adjustment. This construction yields 91 speaker clusters on CREMA-D and 10 on
VESUS; the latter necessarily produces less precise condition-level
intervals.

\section{Endpoint Audit}
\label{app:endpoint}

Table~\ref{tab:app-endpoint} reports three implementation checks for the answer endpoint.

\paragraph{Answer surfaces.}
For each system we enumerate candidate surfaces with its own tokenizer and select the one whose tokens carry the full-vocabulary top-1 mass. Every selected option verbalizer is a single native vocabulary token, so its logit is obtained from one output-head row rather than an aggregation across subtokens. The choice of surface is not cosmetic. Audio-Flamingo-3 puts all of its mass on parenthesis-merged tokens and none on bare letters, so an analysis that assumed bare letters would have scored a token that system never emits. Phi-4 places 0.9998 and 0.9992 of its top-1 mass on bare letters, with the small remainder on the parenthesis variant. Spaced surfaces receive no mass in any condition.

\paragraph{Off-format generations.}
The strict parser accepts only a leading option marker. Under the uncorrected
generation protocol, all generated answers satisfy this requirement.

\paragraph{State-to-logit reconstruction and conditioning.}
Passing each saved post-normalization state back through the native language-model head reproduces the stored option-logit contrasts with a mean absolute error one to two orders of magnitude below the tolerance derived from each condition's logit scale. The option-contrast matrix is well conditioned everywhere, with a condition number between 2.51 and 2.94 for four systems and 8.35 for Phi-4.

\begin{table}[t]
\scriptsize
\centering
\setlength{\tabcolsep}{3pt}
\caption{Endpoint audit. \emph{Surface} is the selected answer surface and its full-vocabulary top-1 in-option rate; the alternative surface is given where the sweep was stored. \emph{Off-fmt} is the fraction of generations the strict parser rejected. \emph{MAE/tol} is the state-to-logit reconstruction error against its tolerance, and \(\kappa\) is the condition number of the option-contrast matrix.}
\label{tab:app-endpoint}
\begin{tabular}{llllrrl}
\toprule
System & Corpus & Surface & In-option & Off-fmt & \(\kappa\) & MAE/tol \\
\midrule
AF3       & CREMA-D & paren & 1.0000 & 0.0000 & 2.94 & 0.026/0.425 \\
AF3       & VESUS   & paren & 1.0000 & 0.0000 & 2.94 & 0.026/0.420 \\
Kimi      & CREMA-D & bare  & 1.0000 & 0.0000 & 2.51 & 0.026/0.363 \\
Kimi      & VESUS   & bare  & 1.0000 & 0.0000 & 2.51 & 0.026/0.368 \\
Phi-4     & CREMA-D & bare  & \phantom{0}0.9998 & 0.0000 & 8.35 & 0.053/0.875 \\
Phi-4     & VESUS   & bare  & \phantom{0}0.9992 & 0.0000 & 8.35 & 0.054/0.825 \\
Omni      & CREMA-D & bare  & --- & 0.0000 & 2.51 & 0.002/0.280 \\
Omni      & VESUS   & bare  & --- & 0.0000 & 2.51 & 0.002/0.284 \\
Q2A       & CREMA-D & bare  & --- & 0.0000 & 2.60 & 0.002/0.265 \\
Q2A       & VESUS   & bare  & --- & 0.0000 & 2.60 & 0.002/0.266 \\
\bottomrule
\end{tabular}
\end{table}

For Audio-Flamingo-3, the parenthesized form is the valid answer surface; the
bare and spaced forms are not selected. The surface sweep was not stored for
the four Qwen conditions, which were run on an earlier pass of the pipeline,
but all of their generated answers pass the strict parser and use one of the
requested option forms.

\section{The Performance Ladder in Full}
\label{app:ladder-full}

Table~\ref{tab:app-ladder-full} gives every rung and both explanatory gaps with speaker-clustered intervals for all 10 conditions. \(A_{\mathrm{MLP}}\) is the nonlinear control of Section~\ref{app:robust}: a multilayer perceptron on the same three option contrasts.

\begin{table*}[t]
\scriptsize
\centering
\setlength{\tabcolsep}{5pt}
\caption{The full ladder, ordered by \(A_{\mathrm{gen}}\). Intervals are paired speaker-clustered bootstraps and are not family-wise adjusted.}
\label{tab:app-ladder-full}
\begin{tabular}{llrrrrrrll}
\toprule
System & Corpus & \(A_{\mathrm{gen}}\) & \(A_{\mathrm{opt}}\) & \(A_{\mathrm{aff}}\) & \(A_{\mathrm{MLP}}\) & \(A_{\mathrm{state}}\) & \(\Delta_{\mathrm{endpoint}}\) & \(\Delta_{\mathrm{decision}}\) [95\% CI] & \(\Delta_{\mathrm{coverage}}\) [95\% CI] \\
\midrule
AF3       & CREMA-D  & 0.9060 & 0.9060 & 0.9426 & 0.9434 & 0.9487 & 0.0000 & $+0.0366$\,[$+0.0289,+0.0446$] & $+0.0062$\,[$+0.0022,+0.0104$] \\
Q2A       & CREMA-D  & 0.7025 & 0.7025 & 0.8995 & 0.9033 & 0.9594 & 0.0000 & $+0.1970$\,[$+0.1909,+0.2030$] & $+0.0599$\,[$+0.0540,+0.0658$] \\
AF3       & VESUS    & 0.5667 & 0.5667 & 0.6335 & 0.6257 & 0.6955 & 0.0000 & $+0.0668$\,[$+0.0244,+0.1065$] & $+0.0620$\,[$+0.0122,+0.1037$] \\
Omni      & CREMA-D  & 0.5265 & 0.5265 & 0.7332 & 0.7448 & 0.8714 & 0.0000 & $+0.2067$\,[$+0.1933,+0.2190$] & $+0.1382$\,[$+0.1275,+0.1492$] \\
Kimi      & CREMA-D  & 0.3994 & 0.3994 & 0.5852 & 0.6126 & 0.8572 & 0.0000 & $+0.1858$\,[$+0.1741,+0.1977$] & $+0.2720$\,[$+0.2577,+0.2857$] \\
Kimi      & VESUS    & 0.3704 & 0.3704 & 0.4104 & 0.4230 & 0.6684 & 0.0000 & $+0.0400$\,[$+0.0263,+0.0530$] & $+0.2581$\,[$+0.2277,+0.2921$] \\
Q2A       & VESUS    & 0.3677 & 0.3677 & 0.4358 & 0.4425 & 0.6029 & 0.0000 & $+0.0681$\,[$+0.0501,+0.0888$] & $+0.1671$\,[$+0.1451,+0.1909$] \\
Omni      & VESUS    & 0.3590 & 0.3590 & 0.4302 & 0.4385 & 0.6290 & 0.0000 & $+0.0712$\,[$+0.0508,+0.0917$] & $+0.1988$\,[$+0.1635,+0.2345$] \\
Phi-4     & CREMA-D  & 0.2793 & 0.2793 & 0.3544 & 0.3759 & 0.7223 & 0.0000 & $+0.0751$\,[$+0.0685,+0.0820$] & $+0.3679$\,[$+0.3508,+0.3838$] \\
Phi-4     & VESUS    & 0.2594 & 0.2595 & 0.2859 & 0.2949 & 0.5660 & 0.0000 & $+0.0264$\,[$+0.0130,+0.0401$] & $+0.2801$\,[$+0.2419,+0.3250$] \\
\bottomrule
\end{tabular}
\end{table*}

\section{Robustness of the Readout-Coverage Gap}
\label{app:robust}

Three controls target three different alternatives to the coverage gap. Table~\ref{tab:app-robust} reports the second and third; the first is the \(A_{\mathrm{MLP}}\) column of Table~\ref{tab:app-ladder-full}.

\paragraph{The affine function class is not the bottleneck.}
Replacing the affine reader on the three option contrasts with a multilayer perceptron on the same contrasts does not absorb the state advantage. \(A_{\mathrm{state}}-A_{\mathrm{MLP}}\) ranges from \(+0.0053\) to \(+0.3464\), and its interval excludes zero in all 10 conditions. The nonlinear reader beats the affine one on the same contrasts in nine conditions, so the option scores do carry some nonlinearly accessible structure; it is simply far smaller than what the full state supports.

\paragraph{The selection budget is not the explanation.}
\(A_{\mathrm{aff}}\) selects its penalty by nested cross-validation while \(A_{\mathrm{state}}\) uses a fixed penalty, so the subtracted term is the tuned one. Refitting both ends under a matched nested budget changes the coverage gap by at most \(0.0164\) in absolute accuracy (Qwen2.5-Omni \(\times\) VESUS) and leaves five conditions slightly lower than reported. Every coverage interval still excludes zero. The main text reports the fixed-penalty specification; the matched refit confirms the result without systematically favoring either reader.

\paragraph{No single speaker drives either gap.}
Recomputing both terms with each speaker's rows removed in turn, without refitting, leaves both terms positive in every replicate of every condition. This check does not use the bootstrap's resampling assumptions at all, which matters most for the 10-speaker corpus.

\begin{table*}[t]
\scriptsize
\centering
\setlength{\tabcolsep}{5pt}
\caption{Capacity-matched refit and delete-one-speaker jackknife. \(\Delta\) is the change in the coverage gap under the matched budget relative to Table~\ref{tab:app-ladder-full}.}
\label{tab:app-robust}
\begin{tabular}{llrlrll}
\toprule
& & \multicolumn{3}{c}{Capacity-matched refit} & \multicolumn{2}{c}{Jackknife range} \\
\cmidrule(lr){3-5}\cmidrule(lr){6-7}
System & Corpus & \(A_{\mathrm{state}}\) & \(\Delta_{\mathrm{coverage}}\) [95\% CI] & \(\Delta\) & \(\Delta_{\mathrm{decision}}\) & \(\Delta_{\mathrm{coverage}}\) \\
\midrule
AF3       & CREMA-D  & 0.9487 & $+0.0062$\,[$+0.0022,+0.0104$] & $+0.0000$ & [$+0.0348,+0.0372$] & [$+0.0057,+0.0068$] \\
Q2A       & CREMA-D  & 0.9592 & $+0.0597$\,[$+0.0538,+0.0657$] & $-0.0002$ & [$+0.1961,+0.1980$] & [$+0.0589,+0.0606$] \\
AF3       & VESUS    & 0.6911 & $+0.0576$\,[$+0.0045,+0.1017$] & $-0.0044$ & [$+0.0546,+0.0789$] & [$+0.0527,+0.0800$] \\
Omni      & CREMA-D  & 0.8715 & $+0.1383$\,[$+0.1273,+0.1495$] & $+0.0001$ & [$+0.2051,+0.2086$] & [$+0.1369,+0.1402$] \\
Kimi      & CREMA-D  & 0.8562 & $+0.2710$\,[$+0.2569,+0.2846$] & $-0.0010$ & [$+0.1845,+0.1876$] & [$+0.2703,+0.2751$] \\
Kimi      & VESUS    & 0.6760 & $+0.2657$\,[$+0.2338,+0.3047$] & $+0.0076$ & [$+0.0365,+0.0449$] & [$+0.2458,+0.2662$] \\
Q2A       & VESUS    & 0.6034 & $+0.1676$\,[$+0.1447,+0.1923$] & $+0.0005$ & [$+0.0602,+0.0720$] & [$+0.1586,+0.1737$] \\
Omni      & VESUS    & 0.6455 & $+0.2153$\,[$+0.1787,+0.2506$] & $+0.0164$ & [$+0.0659,+0.0769$] & [$+0.1876,+0.2100$] \\
Phi-4     & CREMA-D  & 0.7221 & $+0.3678$\,[$+0.3507,+0.3837$] & $-0.0002$ & [$+0.0739,+0.0760$] & [$+0.3656,+0.3702$] \\
Phi-4     & VESUS    & 0.5659 & $+0.2800$\,[$+0.2420,+0.3248$] & $-0.0001$ & [$+0.0230,+0.0296$] & [$+0.2628,+0.2907$] \\
\bottomrule
\end{tabular}
\end{table*}

\section{Logit Correction Details}
\label{app:offset}

Offsets are estimated per condition and per prompt variant from the marginal option probabilities of the evaluation rows, using no emotion labels, and are centered before use. For every condition, the offsets and the corrected accuracy they imply were written to disk before the corrected-generation pass ran, so each row of Table~\ref{tab:app-offset} compares a generated result against a prediction fixed in advance. The four Qwen conditions were re-estimated on the full corpora for this table; their baselines reproduce \(A_{\mathrm{gen}}\) to four decimals.

Prediction and generation agree exactly in five conditions. Qwen2.5-Omni
\(\times\) VESUS is the only condition in which correction reduces
parseability: 0.0019 of rows leave the option set. The remaining disagreements
change one valid option into another and therefore do not create a format cost.

\begin{table*}[t]
\scriptsize
\centering
\setlength{\tabcolsep}{5pt}
\caption{Label-free logit correction. \emph{Predicted} is the corrected accuracy registered before the generation pass; \emph{generated} is the observed one. \emph{Flip} gives the predicted and observed fraction of rows whose answer changes. \emph{Recovery} is the gain divided by \(\Delta_{\mathrm{decision}}\).}
\label{tab:app-offset}
\begin{tabular}{llrrrrrlrr}
\toprule
System & Corpus & Baseline & Predicted & Generated & Gain & Agreement & Flip pred./obs. & Parseability & Recovery \\
\midrule
AF3       & CREMA-D & 0.9060 & 0.9158 & 0.9158 & $+0.0098$ & 1.0000 & 0.0131 / 0.0131 & 1.000 $\to$ 1.000 & 0.268 \\
Q2A       & CREMA-D & 0.7025 & 0.8177 & 0.8177 & $+0.1152$ & 1.0000 & 0.1401 / 0.1401 & 1.000 $\to$ 1.000 & 0.585 \\
AF3       & VESUS   & 0.5667 & 0.6137 & 0.6137 & $+0.0470$ & 1.0000 & 0.1155 / 0.1155 & 1.000 $\to$ 1.000 & 0.703 \\
Omni      & CREMA-D & 0.5265 & 0.6657 & 0.6657 & $+0.1392$ & 1.0000 & 0.2137 / 0.2137 & 1.000 $\to$ 1.000 & 0.674 \\
Kimi      & CREMA-D & 0.3994 & 0.5343 & 0.5289 & $+0.1295$ & \phantom{0}0.9685 & 0.3833 / 0.3607 & 1.000 $\to$ 1.000 & 0.697 \\
Kimi      & VESUS   & 0.3704 & 0.3920 & 0.3898 & $+0.0194$ & \phantom{0}0.9553 & 0.3025 / 0.2947 & 1.000 $\to$ 1.000 & 0.485 \\
Q2A       & VESUS   & 0.3677 & 0.4268 & 0.4268 & $+0.0591$ & 1.0000 & 0.3616 / 0.3616 & 1.000 $\to$ 1.000 & 0.868 \\
Omni      & VESUS   & 0.3590 & 0.4043 & 0.4036 & $+0.0446$ & \phantom{0}0.9981 & 0.2008 / 0.2007 & 1.000 $\to$ \phantom{0}0.998 & 0.627 \\
Phi-4     & CREMA-D & 0.2793 & 0.3193 & 0.3192 & $+0.0399$ & \phantom{0}0.9993 & 0.3506 / 0.3501 & 1.000 $\to$ 1.000 & 0.532 \\
Phi-4     & VESUS   & 0.2594 & 0.2772 & 0.2772 & $+0.0178$ & \phantom{0}0.9995 & 0.2791 / 0.2787 & 1.000 $\to$ 1.000 & 0.674 \\
\bottomrule
\end{tabular}
\end{table*}

The recovery ratio is reported per condition rather than summarized, because it is unstable when its denominator is small and because the supervised affine reader is a diagnostic reference rather than a target the label-free offset is expected to reach. The main text quotes the 0.27 to 0.87 range across all 10 conditions.

\section{Readout Subspaces and Held-Out Decoding}
\label{app:subspaces}

\begin{table}[t]
\scriptsize
\centering
\setlength{\tabcolsep}{4pt}
\caption{Held-out rank-three subspace decodability at \(L^*\), plotted in main-text Fig.~\ref{fig:availuse}(a). \(\mathrm{rand}_3\) is averaged over 20 independent draws. Intervals for \(V_3-\mathrm{rand}_3\) are speaker-clustered over the held-out speakers and exclude zero except on Phi-4-MM \(\times\) VESUS.}
\label{tab:subspace}
\setlength{\tabcolsep}{2pt}
\resizebox{\columnwidth}{!}{%
\begin{tabular}{llrrlr}
\toprule
System & Corpus & \(V_3\) & \(\mathrm{rand}_3\) &
\(V_3-\mathrm{rand}_3\) [95\% CI] & \(S_{\mathrm{decoding}}\) \\
\midrule
Qwen2-Audio  & CREMA-D & 0.890 & 0.618 & $+0.272$\,[$+0.260,+0.284$] & 0.949 \\
Qwen2.5-Omni & CREMA-D & 0.732 & 0.528 & $+0.204$\,[$+0.190,+0.218$] & 0.858 \\
Audio-Flamingo-3 & CREMA-D & 0.937 & 0.869 & $+0.068$\,[$+0.060,+0.077$] & 0.940 \\
Kimi-Audio   & CREMA-D & 0.602 & 0.467 & $+0.136$\,[$+0.117,+0.155$] & 0.823 \\
Phi-4-MM     & CREMA-D & 0.342 & 0.320 & $+0.022$\,[$+0.004,+0.040$] & 0.668 \\
Qwen2-Audio  & VESUS   & 0.463 & 0.359 & $+0.104$\,[$+0.075,+0.132$] & 0.606 \\
Qwen2.5-Omni & VESUS   & 0.464 & 0.351 & $+0.112$\,[$+0.062,+0.151$] & 0.561 \\
Audio-Flamingo-3 & VESUS & 0.670 & 0.561 & $+0.109$\,[$+0.082,+0.135$] & 0.690 \\
Kimi-Audio   & VESUS   & 0.408 & 0.372 & $+0.037$\,[$+0.026,+0.047$] & 0.563 \\
Phi-4-MM     & VESUS   & 0.259 & 0.277 & $-0.018$\,[$-0.033,+0.001$] & 0.481 \\
\bottomrule
\end{tabular}
}
\end{table}

With seed 0, half of the speakers are assigned to the training split and the
remainder to the held-out split; construction is prompt-specific. Let
\(W_{\mathrm{opt}}\) stack the four option-token output rows,
\(V_4=\operatorname{row}(W_{\mathrm{opt}})\), and \(P_{V_4}\) project onto
that space. Let \(H\) contain training states at \(L^*\), and let
\(\mathcal{N}_f\) be the final normalization. We standardize
\(H_{\perp}=\mathcal{N}_f(H)(I-P_{V_4})\), fit an L2 multinomial logistic
regression (\(C=0.01\), 3{,}000 iterations), and define
\begin{equation}
\begin{aligned}
B_{\perp}
  &=B_{\mathrm{std}}D^{-1}(I-P_{V_4})
    =U\Sigma R^\top,\\
S_{\mathrm{decoding}}
  &=\operatorname{span}\{R_{i,:}\}_{i=1}^{3},
\end{aligned}
\label{eq:app-sperp}
\end{equation}
where \(B_{\mathrm{std}}\) is the standardized coefficient matrix and \(D\)
contains the fitted feature scales. Reprojection and re-orthonormalization
reduce leakage into \(V_4\) below \(10^{-6}\). Because the relative-contrast
space \(V_3\) is contained in \(V_4\),
\(S_{\mathrm{decoding}}\subset V_4^\perp\subset V_3^\perp\).

The rank-four space \(S_{\mathrm{intervention}}\) used in the causal analysis is built from
the leading four discriminant directions. Because both spaces come from the
same singular basis, \(S_{\mathrm{decoding}}\) is contained in \(S_{\mathrm{intervention}}\) by
construction. The rank-four intervention therefore contains the decoded
directions, although causal effects need not be monotone under subspace
expansion. The causal analysis uses the matched rank-four geometry.

Held-out decodability uses a separate standardized L2 multinomial probe
(\(C=0.5\), 2{,}000 iterations), fitted on the training split and scored on
the held-out split. Correctness is averaged over four prompt variants,
and 95\% intervals use 2{,}000 speaker-bootstrap resamples. For example,
Phi-4 \(\times\) CREMA-D uses 986 clips from 45 training speakers and 1{,}014
clips from 46 held-out speakers. Each \(\mathrm{rand}_3\) draw is a
standard-normal sample projected into \(V_4^\perp\) and
re-orthonormalized, with the raw seed reused across the four prompts so that only
the per-prompt projector differs.

\subsection{Averaging the random reference over draws}
\label{app:rand3-draws}

A single random subspace is a noisy reference. Repeating the entire probe over 20
independent draws (Table~\ref{tab:app-rand3draws}) shows that same-rank draws vary
by \(0.017\) to \(0.053\) in standard deviation, enough to move a small contrast
across zero. The decoding results in the main text therefore use the draw-averaged
reference: per-clip correctness is averaged over the 20 draws before scoring and
bootstrapping, exactly as it is averaged over the four prompts.

The native contrast is positive on all 20 draws in seven conditions. It is
positive on 16 draws for Kimi-Audio \(\times\) VESUS, 17 for Phi-4-MM \(\times\)
CREMA-D, and only 2 for Phi-4-MM \(\times\) VESUS, confirming that small native
contrasts can depend on the random draw. By contrast,
\(S_{\mathrm{decoding}}-\mathrm{rand}_3\) is positive on all 20 draws in all ten conditions.

\begin{table}[t]
\scriptsize
\centering
\setlength{\tabcolsep}{3pt}
\caption{The random reference across 20 independent rank-three draws. \emph{sd} and \emph{range} are over draws, not over speakers. The last two columns count the draws on which each contrast is positive.}
\label{tab:app-rand3draws}
\setlength{\tabcolsep}{2pt}
\resizebox{\columnwidth}{!}{%
\begin{tabular}{llrrlcc}
\toprule
System & Corpus & mean & sd & range & \(V_3{>}\mathrm{rand}_3\) & \(S_{\mathrm{decoding}}{>}\mathrm{rand}_3\) \\
\midrule
Qwen2-Audio  & CREMA-D & 0.618 & 0.053 & [0.506,0.719] & 20/20 & 20/20 \\
Qwen2.5-Omni & CREMA-D & 0.528 & 0.030 & [0.479,0.583] & 20/20 & 20/20 \\
Audio-Flamingo-3 & CREMA-D & 0.869 & 0.024 & [0.831,0.908] & 20/20 & 20/20 \\
Kimi-Audio   & CREMA-D & 0.467 & 0.030 & [0.399,0.521] & 20/20 & 20/20 \\
Phi-4-MM     & CREMA-D & 0.320 & 0.019 & [0.296,0.361] & 17/20 & 20/20 \\
Qwen2-Audio  & VESUS   & 0.359 & 0.022 & [0.315,0.394] & 20/20 & 20/20 \\
Qwen2.5-Omni & VESUS   & 0.351 & 0.021 & [0.317,0.392] & 20/20 & 20/20 \\
Audio-Flamingo-3 & VESUS & 0.561 & 0.027 & [0.511,0.607] & 20/20 & 20/20 \\
Kimi-Audio   & VESUS   & 0.372 & 0.034 & [0.305,0.429] & 16/20 & 20/20 \\
Phi-4-MM     & VESUS   & 0.277 & 0.017 & [0.249,0.311] & \phantom{0}2/20 & 20/20 \\
\bottomrule
\end{tabular}
}
\end{table}

Table~\ref{tab:app-subspace} adds the intervals omitted from the main text. \(S_{\mathrm{decoding}}-\mathrm{rand}_3\) is positive with an interval excluding zero in all ten conditions. Adding \(S_{\mathrm{decoding}}\) to \(V_3\) also improves decoding in eight conditions; the two nonsignificant Audio-Flamingo-3 increments reflect that its native readout is already highly informative.

\begin{table*}[t]
\scriptsize
\centering
\setlength{\tabcolsep}{4pt}
\caption{Held-out subspace decodability with intervals. \(V_3-\mathrm{rand}_3\) compares two spaces that are both fixed without emotion labels; \(S_{\mathrm{decoding}}\) is selected on the training split, so its margin over \(\mathrm{rand}_3\) includes a supervised search advantage. \emph{n.s.} marks an interval containing zero.}
\label{tab:app-subspace}
\begin{tabular}{llrlll}
\toprule
System & Corpus & \(S_{\mathrm{decoding}}\) & \(V_3-\mathrm{rand}_3\) & \(S_{\mathrm{decoding}}-\mathrm{rand}_3\) & \([V_3,S_{\mathrm{decoding}}]-V_3\) \\
\midrule
Qwen2-Audio  & CREMA-D & 0.949 & $+0.272$\,[$+0.260,+0.284$] & $+0.332$\,[$+0.321,+0.342$] & $+0.059$\,[$+0.048,+0.070$] \\
Qwen2.5-Omni & CREMA-D & 0.858 & $+0.204$\,[$+0.190,+0.218$] & $+0.330$\,[$+0.309,+0.350$] & $+0.125$\,[$+0.103,+0.147$] \\
Audio-Flamingo-3 & CREMA-D & 0.940 & $+0.068$\,[$+0.060,+0.077$] & $+0.071$\,[$+0.062,+0.080$] & $+0.003$\,[$-0.003,+0.011$] n.s. \\
Kimi-Audio   & CREMA-D & 0.823 & $+0.136$\,[$+0.117,+0.155$] & $+0.357$\,[$+0.325,+0.386$] & $+0.221$\,[$+0.185,+0.255$] \\
Phi-4-MM     & CREMA-D & 0.668 & $+0.022$\,[$+0.004,+0.040$] & $+0.348$\,[$+0.320,+0.375$] & $+0.328$\,[$+0.294,+0.361$] \\
Qwen2-Audio  & VESUS   & 0.606 & $+0.104$\,[$+0.075,+0.132$] & $+0.246$\,[$+0.179,+0.322$] & $+0.144$\,[$+0.103,+0.199$] \\
Qwen2.5-Omni & VESUS   & 0.561 & $+0.112$\,[$+0.062,+0.151$] & $+0.209$\,[$+0.155,+0.257$] & $+0.099$\,[$+0.077,+0.128$] \\
Audio-Flamingo-3 & VESUS & 0.690 & $+0.109$\,[$+0.082,+0.135$] & $+0.129$\,[$+0.105,+0.160$] & $+0.017$\,[$-0.032,+0.070$] n.s. \\
Kimi-Audio   & VESUS   & 0.563 & $+0.037$\,[$+0.026,+0.047$] & $+0.192$\,[$+0.140,+0.241$] & $+0.156$\,[$+0.110,+0.201$] \\
Phi-4-MM     & VESUS   & 0.481 & $-0.018$\,[$-0.033,+0.001$] n.s. & $+0.204$\,[$+0.144,+0.287$] & $+0.226$\,[$+0.175,+0.294$] \\
\bottomrule
\end{tabular}
\end{table*}

\section{Minimal-Pair Activation Replacement}
\label{app:patching}

\subsection{Pairing, per-arm rates, and statistical units}

The primary causal pass uses strict minimal pairs from the held-out split: receiver and donor share speaker and transcript but differ in emotion. The donor representation is the state of one real clip under the same prompt, not a training-set or class-average state, and the same clip pair is used across all arms and prompts. The training split still fixes \(L^*\), \(S_{\mathrm{intervention}}\), and the random subspace before any held-out intervention.

Within the 2{,}000-clip stratified sample, 866 of 1{,}014 held-out CREMA-D clips and 474 of 1{,}000 held-out VESUS clips have an eligible minimal-pair donor. We sample 400 eligible receivers per condition and evaluate four prompts, giving 1{,}600 rows in each of the ten conditions. Because receiver and donor share a speaker, intervals cluster the paired outcomes by speaker: 46 clusters on CREMA-D and 5 on VESUS.

Table~\ref{tab:patching} reports the primary per-condition effects that main-text Fig.~\ref{fig:availuse}(b) plots, together with each condition's \(L^*\).

\begin{table*}[t]
\scriptsize
\centering
\setlength{\tabcolsep}{6pt}
\caption{Matched rank-four activation replacement at \(L^*\) using held-out minimal pairs. The in-span arm replaces \(V_4\) and the readout-external arm replaces \(S_{\mathrm{intervention}}\); each effect is the paired difference in answer-change rate from the same \(\mathrm{rand}_4\) arm, with speaker-clustered 95\% intervals, and n.s.\ marks an interval containing zero. The last column places the readout-external effect on a common scale as its one-sided 95\% upper bound divided by the in-span effect of the same condition.}
\label{tab:patching}
\begin{tabular}{llrllr}
\toprule
System & Corpus & \(L^*\) & In-span effect [95\% CI] & Readout-external effect [95\% CI] & Relative 95\% UB \\
\midrule
Qwen2-Audio  & CREMA-D & 28 & $+0.3344$\,[$+0.3115,+0.3572$] & $+0.0206$\,[$+0.0136,+0.0276$]      & 7.9\% \\
Qwen2.5-Omni & CREMA-D & 27 & $+0.3631$\,[$+0.3218,+0.4044$] & $+0.0037$\,[$+0.0005,+0.0070$]      & 1.8\% \\
Audio-Flamingo-3 & CREMA-D & 25 & $+0.1156$\,[$+0.0915,+0.1397$] & $+0.0025$\,[$+0.0002,+0.0048$]      & 3.8\% \\
Kimi-Audio   & CREMA-D & 27 & $+0.1950$\,[$+0.1582,+0.2318$] & $+0.0050$\,[$+0.0019,+0.0081$]      & 3.9\% \\
Phi-4-MM     & CREMA-D & 29 & $+0.0606$\,[$+0.0465,+0.0747$] & $-0.0063$\,[$-0.0127,+0.0002$] n.s. & $-1.4\%$ \\
Qwen2-Audio  & VESUS   & 31 & $+0.2437$\,[$+0.1617,+0.3258$] & $+0.0000$\,[$-0.0017,+0.0017$] n.s. & 0.6\% \\
Qwen2.5-Omni & VESUS   & 27 & $+0.1994$\,[$+0.1110,+0.2877$] & $+0.0000$\,[$-0.0017,+0.0017$] n.s. & 0.7\% \\
Audio-Flamingo-3 & VESUS   & 25 & $+0.1319$\,[$+0.0839,+0.1798$] & $-0.0013$\,[$-0.0054,+0.0029$] n.s. & 1.7\% \\
Kimi-Audio   & VESUS   & 24 & $+0.0181$\,[$+0.0119,+0.0243$] & $+0.0088$\,[$+0.0053,+0.0122$]      & 64.1\% \\
Phi-4-MM     & VESUS   & 28 & $+0.0269$\,[$+0.0185,+0.0352$] & $-0.0006$\,[$-0.0058,+0.0046$] n.s. & 13.9\% \\
\bottomrule
\end{tabular}
\end{table*}

\begin{table*}[t]
\scriptsize
\centering
\setlength{\tabcolsep}{5pt}
\caption{Minimal-pair answer-change rates by arm at \(L^*\). \(V\), \(S_{\mathrm{intervention}}\), and Random are rank matched; Full replaces the entire state.}
\label{tab:app-patch-arms}
\begin{tabular}{llrrrr}
\toprule
System & Corpus & \(V\) & \(S_{\mathrm{intervention}}\) & Random & Full \\
\midrule
Q2A   & CREMA-D & 0.3350 & 0.0213 & 0.0006 & 0.6275 \\
Omni  & CREMA-D & 0.3631 & 0.0037 & 0.0000 & 0.5081 \\
AF3   & CREMA-D & 0.1175 & 0.0044 & 0.0019 & 0.8831 \\
Kimi  & CREMA-D & 0.1963 & 0.0063 & 0.0013 & 0.3550 \\
Phi-4 & CREMA-D & 0.0744 & 0.0075 & 0.0138 & 0.2744 \\
Q2A   & VESUS   & 0.2450 & 0.0013 & 0.0013 & 0.2881 \\
Omni  & VESUS   & 0.2000 & 0.0006 & 0.0006 & 0.2762 \\
AF3   & VESUS   & 0.1350 & 0.0019 & 0.0031 & 0.5713 \\
Kimi  & VESUS   & 0.0300 & 0.0206 & 0.0119 & 0.3144 \\
Phi-4 & VESUS   & 0.0413 & 0.0138 & 0.0144 & 0.1363 \\
\bottomrule
\end{tabular}
\end{table*}

Subtracting the random rate gives positive in-span effects in all ten conditions, from \(+0.0181\) to \(+0.3631\). The readout-external effect reaches \(+0.0206\) at most, and its relative upper bound is at most 7.9\% in eight conditions. The two larger ratios occur where the in-span denominator is small; Kimi-Audio \(\times\) VESUS is the only one of them with a detected readout-external answer-change effect.

\subsection{Donor-content outcome}

Answer change asks whether an edit moves the answer; donor following asks whether it moves specifically toward the donor emotion. Table~\ref{tab:app-patch-follow} reports both matched effects on this stricter outcome.

\begin{table*}[t]
\scriptsize
\centering
\setlength{\tabcolsep}{7pt}
\caption{Minimal-pair donor-following effects relative to the same random arm, with speaker-clustered 95\% intervals.}
\label{tab:app-patch-follow}
\begin{tabular}{llll}
\toprule
System & Corpus & In-span effect [95\% CI] & Readout-external effect [95\% CI] \\
\midrule
Q2A   & CREMA-D & $+0.2281$\,[$+0.2056,+0.2506$] & $+0.0106$\,[$+0.0058,+0.0154$] \\
Omni  & CREMA-D & $+0.2400$\,[$+0.2025,+0.2775$] & $+0.0006$\,[$-0.0006,+0.0018$] n.s. \\
AF3   & CREMA-D & $+0.1088$\,[$+0.0857,+0.1318$] & $+0.0013$\,[$-0.0005,+0.0030$] n.s. \\
Kimi  & CREMA-D & $+0.1181$\,[$+0.0894,+0.1468$] & $+0.0006$\,[$-0.0015,+0.0027$] n.s. \\
Phi-4 & CREMA-D & $+0.0031$\,[$-0.0072,+0.0134$] n.s. & $-0.0031$\,[$-0.0077,+0.0014$] n.s. \\
Q2A   & VESUS   & $+0.1494$\,[$+0.0969,+0.2019$] & $-0.0006$\,[$-0.0017,+0.0004$] n.s. \\
Omni  & VESUS   & $+0.1156$\,[$+0.0624,+0.1689$] & $+0.0000$\,[$+0.0000,+0.0000$] n.s. \\
AF3   & VESUS   & $+0.0950$\,[$+0.0445,+0.1455$] & $-0.0013$\,[$-0.0050,+0.0025$] n.s. \\
Kimi  & VESUS   & $+0.0106$\,[$+0.0055,+0.0157$] & $+0.0000$\,[$-0.0048,+0.0048$] n.s. \\
Phi-4 & VESUS   & $-0.0050$\,[$-0.0113,+0.0013$] n.s. & $-0.0031$\,[$-0.0064,+0.0001$] n.s. \\
\bottomrule
\end{tabular}
\end{table*}

Only Qwen2-Audio \(\times\) CREMA-D shows readout-external donor following above random. Its \(+0.0106\) effect shows limited content-specific causal influence under a compatible replacement. The Phi-4-MM in-span arm does not transfer donor content in either corpus, so its already small answer-change effects provide a weak scale reference.

\subsection{Perturbation-magnitude diagnostic}

Minimal pairing makes donor and receiver states more similar, so a null effect can coincide with a smaller edit. Table~\ref{tab:app-patch-magnitude} reports the applied relative state change in all ten conditions and, where an earlier cross-speaker pass is available, the minimal-to-cross-speaker ratio.

\begin{table*}[t]
\scriptsize
\centering
\setlength{\tabcolsep}{5pt}
\caption{Mean relative intervention magnitude \(\|\widetilde h_r-h_r\|/\|h_r\|\). Ratios compare minimal-pair with cross-speaker replacement where available; the last column compares the two arms within the minimal-pair pass.}
\label{tab:app-patch-magnitude}
\begin{tabular}{llrrrrr}
\toprule
System & Corpus & Minimal \(V\) & Minimal \(S_{\mathrm{intervention}}\) & \(V\) ratio & \(S_{\mathrm{intervention}}\) ratio & \(S_{\mathrm{intervention}}/V\) \\
\midrule
Q2A   & CREMA-D & 0.0786 & 0.1750 & 0.98 & 0.95 & 2.23 \\
Omni  & CREMA-D & 0.0585 & 0.0233 & 0.91 & 0.93 & 0.40 \\
AF3   & CREMA-D & 0.1327 & 0.0987 & -- & -- & 0.74 \\
Kimi  & CREMA-D & 0.0453 & 0.0203 & -- & -- & 0.45 \\
Phi-4 & CREMA-D & 0.0197 & 0.0226 & 0.71 & 0.88 & 1.15 \\
Q2A   & VESUS   & 0.0248 & 0.0279 & 0.65 & 0.76 & 1.12 \\
Omni  & VESUS   & 0.0296 & 0.0066 & 0.56 & 0.73 & 0.22 \\
AF3   & VESUS   & 0.0818 & 0.0307 & -- & -- & 0.38 \\
Kimi  & VESUS   & 0.0128 & 0.0146 & -- & -- & 1.14 \\
Phi-4 & VESUS   & 0.0126 & 0.0137 & 0.40 & 0.68 & 1.09 \\
\bottomrule
\end{tabular}
\end{table*}

For Qwen2-Audio \(\times\) CREMA-D, minimal pairing retains 95\% of the cross-speaker \(S_{\mathrm{intervention}}\) edit magnitude. Within the minimal-pair pass, that edit is 2.23 times the \(V_4\) edit, yet its answer-change effect is \(+0.0206\) rather than \(+0.3344\), ruling out a weaker external edit as the explanation in this condition. Across all ten conditions, \(S_{\mathrm{intervention}}\) is at least as large as \(V_4\) in five. Kimi-Audio \(\times\) VESUS belongs to this set, but both edits are the smallest in the study, matching its status as a weak-intervention boundary case.

\subsection{Routing capacity above \(L^*\)}
\label{app:routing-capacity}

The selected \(L^*\) lies one to four blocks below the top of each stack (Table~\ref{tab:patching}). A weak \(S_{\mathrm{intervention}}\) effect could therefore have a simple explanation: the remaining blocks might be unable to route any \(V_4^\perp\) component into the option logits.

The full-state arm tests this possibility. Relative to the in-span replacement, it adds the donor's entire \(V_4^\perp\) component. If the remaining blocks were unresponsive to that component, the two arms would have the same effect. Instead, the full arm changes the answer more often than the in-span arm in every condition, from 0.288 versus 0.245 on Qwen2-Audio \(\times\) VESUS to 0.883 versus 0.118 on Audio-Flamingo-3 \(\times\) CREMA-D (Table~\ref{tab:app-patch-arms}).

The additional changes generally move toward the donor emotion. Relative to the unpatched baseline, the full arm raises donor following in all ten conditions: by \(+0.161\) to \(+0.839\) in the eight non-Phi-4-MM conditions, and by \(+0.046\) and \(+0.009\) in the two Phi-4-MM conditions. The weak \(S_{\mathrm{intervention}}\) effects therefore cannot be explained solely by a downstream pathway that is unresponsive to readout-external content.

\subsection{Minimal-pair depth scan: protocol and consistency}
\label{app:minlayer}

% Numbers auto-generated by analyze_patchminlayer.py from
% results/patchminlayer_qwen2audio_cremad_donor_min_L*.npz; formatting adapted to paper notation.
\begin{table*}[t]
\scriptsize
\centering
\setlength{\tabcolsep}{6pt}
\caption{Minimal-pair replacement at eight depths of Qwen2-Audio \(\times\) CREMA-D, on all 866 eligible receivers, with the readout-external subspace refit at each depth. Each cell is the paired difference from the same-rank random arm, with speaker-clustered 95\% intervals; n.s.\ marks an interval containing zero, and \(L=28\) is the \(L^*\) used in Table~\ref{tab:patching}.}
\label{tab:minlayer-depth}
\begin{tabular}{lllll}
\toprule
& \multicolumn{2}{c}{Donor following} & \multicolumn{2}{c}{Answer change} \\
\cmidrule(lr){2-3}\cmidrule(lr){4-5}
\(L\) & In-span \(V_4\) & Readout-external \(S_{\mathrm{intervention}}\) & In-span \(V_4\) & Readout-external \(S_{\mathrm{intervention}}\) \\
\midrule
16 & $+0.0003$\,[$-0.0003,+0.0008$] n.s. & $+0.0012$\,[$+0.0001,+0.0023$] & $+0.0000$\,[$-0.0008,+0.0008$] n.s. & $+0.0014$\,[$-0.0002,+0.0031$] n.s. \\
18 & $+0.0020$\,[$+0.0004,+0.0036$] & $+0.0072$\,[$+0.0039,+0.0105$] & $+0.0032$\,[$+0.0012,+0.0051$] & $+0.0162$\,[$+0.0123,+0.0200$] \\
20 & $+0.0136$\,[$+0.0101,+0.0171$] & $+0.0144$\,[$+0.0099,+0.0190$] & $+0.0185$\,[$+0.0144,+0.0226$] & $+0.0372$\,[$+0.0308,+0.0437$] \\
22 & $+0.0681$\,[$+0.0595,+0.0768$] & $+0.0089$\,[$+0.0058,+0.0121$] & $+0.0941$\,[$+0.0861,+0.1021$] & $+0.0196$\,[$+0.0157,+0.0236$] \\
24 & $+0.0710$\,[$+0.0624,+0.0796$] & $+0.0110$\,[$+0.0073,+0.0147$] & $+0.1028$\,[$+0.0948,+0.1107$] & $+0.0280$\,[$+0.0233,+0.0327$] \\
26 & $+0.1723$\,[$+0.1598,+0.1849$] & $+0.0115$\,[$+0.0083,+0.0148$] & $+0.2610$\,[$+0.2492,+0.2727$] & $+0.0188$\,[$+0.0149,+0.0226$] \\
28 & $+0.2307$\,[$+0.2126,+0.2487$] & $+0.0095$\,[$+0.0062,+0.0128$] & $+0.3346$\,[$+0.3176,+0.3516$] & $+0.0147$\,[$+0.0111,+0.0183$] \\
31 & $+0.4983$\,[$+0.4684,+0.5281$] & $+0.0064$\,[$+0.0036,+0.0091$] & $+0.6544$\,[$+0.6261,+0.6828$] & $+0.0084$\,[$+0.0053,+0.0115$] \\
\bottomrule
\end{tabular}
\end{table*}

Table~\ref{tab:minlayer-depth} repeats the minimal-pair intervention at layers 16, 18, 20, 22, 24, 26, 28, and 31 of Qwen2-Audio \(\times\) CREMA-D. Main-text Fig.~\ref{fig:depth-profile} plots the donor-following columns. The pairing, receiver split, \(V_4\), and \(\mathrm{rand}_4\) are the same as in the primary pass. At each depth, \(S_{\mathrm{intervention}}\) is refit on the training speakers using the construction in Section~\ref{app:subspaces}.

All 866 held-out receivers with an eligible same-speaker, same-transcript donor are included under four prompt variants, giving 3{,}464 rows per cell. Intervals cluster on receiver speaker. At layer 28, the selected \(L^*\), the readout-external answer-change effect is \(+0.0147\)\,[\(+0.0111,+0.0183\)], close to the primary-pass estimate of \(+0.0206\)\,[\(+0.0136,+0.0276\)] obtained from a 400-receiver sample.

Beyond the peak at layer 20, the decline is not layer-by-layer monotone; the effects at layers 22 through 26 sit within one another's intervals, and the ordering claim the table supports is that the external effect is largest at layer 20 and smallest at layers 16 and 31.

\subsection{Propagation of the injected readout-external component}
\label{app:patchtrack}

We repeat the layer-20 and layer-24 \(S_{\mathrm{intervention}}\) replacements on 400 receivers and record the induced answer-position difference
\(\delta(L')=h_{\mathrm{patched}}(L')-h_{\mathrm{unpatched}}(L')\)
at every later layer, together with the endpoint option logits.

After a layer-20 replacement, the component of \(\delta\) in the injected subspace retains 0.88 of its original norm at layer 31. The perturbation therefore persists. Its overlap with \(V_4\) grows from zero at injection to about 0.07 of \(\|\delta\|\) by layer 26 and then remains near that level, showing that part of the perturbation reaches the native readout span.

The endpoint logits also move toward the donor emotion. Relative to the random arm, the donor-option logit minus the mean of the other three option logits shifts by \(+0.190\)\,[\(+0.176,+0.204\)] after the layer-20 replacement and by \(+0.312\)\,[\(+0.289,+0.335\)] after the layer-24 replacement. Thus the injected component reaches the final option scores with the donor's sign, but usually not strongly enough to change their ordering. Because the last saved state may include the model's final normalization, this endpoint statement uses the recorded option logits rather than the state-norm decomposition.

\section{Controls for Measured Surface Acoustic Cues}
\label{app:acoustics}

These analyses test whether the held-out decodability of \(S_{\mathrm{decoding}}\) can be explained by measured surface acoustic cues. We define the descriptor panels, quantify how predictive the cues are, remove them from the subspace coordinates, and separately remove absolute level from the input audio.

\subsection{Descriptor Panel}
\label{app:descriptors}

Ten clip-level acoustic-prosodic descriptors are computed once per recording and are model-independent: duration; voiced-frame fraction; five fundamental-frequency statistics (mean, standard deviation, range, terminal value, and slope of the F0 track); and three RMS-energy statistics (mean, standard deviation, and max-minus-min spread). RMS descriptors enter all emotion-corpus analyses in log units. In the residualization and decoding analyses, missing descriptor values, standardization statistics, ordinary-least-squares coefficients, and class means are all estimated on training-speaker rows only and then applied to held-out rows. The descriptor-only decoder \(A_{\mathrm{desc}}\) uses the same probe family, regularization-selection protocol, and speaker split as the subspace decoders.

An extended panel used for robustness adds ten further descriptors: spectral tilt (the regression slope of the long-term average spectrum in dB over log-frequency), spectral centroid mean and standard deviation, local jitter, local shimmer, harmonics-to-noise ratio, and the first four DCT coefficients of the time-interpolated log-F0 contour. Jitter, shimmer, and harmonics-to-noise ratio are computed with Praat via parselmouth; coverage is complete on both corpora. The nonlinear removal variant replaces the ordinary-least-squares residualization with per-dimension gradient-boosted trees, again fit on training-speaker rows only.

\subsection{Stimulus-Level Loudness Statistics}
\label{app:stimulus-stats}

Table~\ref{tab:app-stimulus} reports, for every pair of emotion classes, the class-mean RMS-level difference in dB and how well dB level alone separates the pair. Folded AUC is \(\max(\mathrm{AUC}, 1-\mathrm{AUC})\) and is therefore orientation-free. Uncertainty is a speaker-clustered bootstrap with 2{,}000 resamples. On CREMA-D, level alone is a strong class separator for several pairs; on VESUS the gaps are smaller. These statistics motivate treating surface loudness as an explicit alternative explanation rather than an afterthought.

\begin{table}[t]
\centering\scriptsize
\setlength{\tabcolsep}{2pt}
\caption{Stimulus-level RMS-loudness differences between emotion classes. \(\Delta\)dB is the class-mean dB RMS gap (first class minus second); folded AUC measures how well dB RMS alone separates the pair. Speaker-clustered bootstrap 95\% intervals.}
\label{tab:app-stimulus}
\begin{tabular}{llll}
\toprule
Corpus & Pair & \(\Delta\)dB [95\% CI] & AUC [95\% CI] \\
\midrule
CREMA-D & happy vs sad & $+8.01\,[+7.48,+8.56]$ & $0.92\,[0.90,0.93]$ \\
 & happy vs angry & $-5.70\,[-6.34,-5.10]$ & $0.77\,[0.75,0.80]$ \\
 & happy vs neutral & $+4.77\,[+4.21,+5.33]$ & $0.80\,[0.76,0.83]$ \\
 & sad vs angry & $-13.71\,[-14.47,-12.96]$ & $0.98\,[0.97,0.99]$ \\
 & sad vs neutral & $-3.24\,[-3.63,-2.85]$ & $0.79\,[0.76,0.82]$ \\
 & angry vs neutral & $+10.47\,[+9.81,+11.15]$ & $0.95\,[0.93,0.96]$ \\
\midrule
VESUS & happy vs sad & $+4.26\,[+2.41,+6.03]$ & $0.70\,[0.62,0.81]$ \\
 & happy vs angry & $-1.23\,[-3.44,+0.74]$ & $0.58\,[0.50,0.71]$ \\
 & happy vs neutral & $+2.56\,[+1.30,+3.98]$ & $0.68\,[0.58,0.81]$ \\
 & sad vs angry & $-5.48\,[-7.67,-3.49]$ & $0.74\,[0.65,0.87]$ \\
 & sad vs neutral & $-1.70\,[-4.04,+0.63]$ & $0.57\,[0.51,0.71]$ \\
 & angry vs neutral & $+3.79\,[+1.52,+5.91]$ & $0.72\,[0.58,0.86]$ \\
\bottomrule
\end{tabular}
\end{table}

\subsection{Base Descriptor Removal}

% Source: phase1_discordance/results/sperp_acoustics_{cell}.json (analyze_sperp_acoustics.py,
% job 26887523; stage-0 gate reproduced tab:subspace exactly, diff 0.0000 in all 10 cells).
\begin{table*}[t]
\scriptsize
\centering
\setlength{\tabcolsep}{6pt}
\caption{Linear controls for the ten measured acoustic-prosodic descriptors at \(L^*\). \(A_{\mathrm{desc}}\) is held-out four-class accuracy from the descriptors alone, with no hidden state. The resid columns re-decode each rank-three projection after removing the descriptors, with the regression fit on training speakers only. The last column is the paired \(S_{\mathrm{decoding}}\) drop with speaker-clustered 95\% intervals; n.s.\ marks an interval containing zero. The \(V_3\) and \(S_{\mathrm{decoding}}\) columns match Table~\ref{tab:subspace}.}
\label{tab:acoustic-char}
\begin{tabular}{llrrrrrl}
\toprule
System & Corpus & \(A_{\mathrm{desc}}\) & \(V_3\) & \(V_3\) resid & \(S_{\mathrm{decoding}}\) & \(S_{\mathrm{decoding}}\) resid & \(S_{\mathrm{decoding}}\) drop [95\% CI] \\
\midrule
Qwen2-Audio  & CREMA-D & 0.640 & 0.890 & 0.759 & 0.949 & 0.856 & $+0.094$\,[$+0.075,+0.115$] \\
Qwen2.5-Omni & CREMA-D & 0.640 & 0.732 & 0.613 & 0.858 & 0.655 & $+0.203$\,[$+0.176,+0.230$] \\
Audio-Flamingo-3 & CREMA-D & 0.640 & 0.937 & 0.852 & 0.940 & 0.847 & $+0.093$\,[$+0.067,+0.121$] \\
Kimi-Audio   & CREMA-D & 0.640 & 0.602 & 0.464 & 0.823 & 0.610 & $+0.214$\,[$+0.189,+0.237$] \\
Phi-4-MM     & CREMA-D & 0.640 & 0.342 & 0.314 & 0.668 & 0.413 & $+0.255$\,[$+0.230,+0.280$] \\
Qwen2-Audio  & VESUS   & 0.343 & 0.463 & 0.426 & 0.606 & 0.537 & $+0.069$\,[$+0.048,+0.089$] \\
Qwen2.5-Omni & VESUS   & 0.343 & 0.464 & 0.422 & 0.561 & 0.493 & $+0.068$\,[$+0.042,+0.095$] \\
Audio-Flamingo-3 & VESUS & 0.343 & 0.670 & 0.644 & 0.690 & 0.651 & $+0.039$\,[$-0.014,+0.077$] n.s. \\
Kimi-Audio   & VESUS   & 0.343 & 0.408 & 0.361 & 0.563 & 0.488 & $+0.076$\,[$+0.040,+0.111$] \\
Phi-4-MM     & VESUS   & 0.343 & 0.259 & 0.262 & 0.481 & 0.423 & $+0.059$\,[$+0.022,+0.090$] \\
\bottomrule
\end{tabular}
\end{table*}

Table~\ref{tab:acoustic-char} reports the base control summarized in main-text Fig.~\ref{fig:acoustic-controls}. It gives the descriptor-only reference \(A_{\mathrm{desc}}\), decoding after linear removal of the ten descriptors, and the paired drop in \(S_{\mathrm{decoding}}\) accuracy.

\subsection{Subspace--Descriptor Associations}
\label{app:desc-assoc-sec}

Table~\ref{tab:app-desc-assoc} reports, for each condition and subspace, the single descriptor best predicted from the rank-three projection, as out-of-sample \(R^2\) under an ordinary-least-squares fit on training speakers. The pooled column predicts the raw descriptor; the within-class column first centers both the descriptor and the projection by their training-speaker class means, so it measures covariation that is not explained by class membership. The dominant descriptors are energy statistics in nearly every condition, and the within-class values are substantially smaller than the pooled ones, indicating that much of the pooled association reflects class structure. These associations are reported descriptively, without a multiplicity correction across descriptors and conditions.

\begin{table*}[t]
\centering\small
\setlength{\tabcolsep}{4pt}
\caption{Strongest acoustic-descriptor association per subspace: the top-\(R^2\) descriptor, with held-out OLS \(R^2\) in parentheses. Pooled: descriptor predicted directly from the rank-three projection. Within-class: both centered by training-speaker class means first.}
\label{tab:app-desc-assoc}
\begin{tabular}{lllll}
\toprule
Condition & \(V_3\) top (pooled) & \(V_3\) top (within) & \(S_{\mathrm{decoding}}\) top (pooled) & \(S_{\mathrm{decoding}}\) top (within) \\
\midrule
Qwen2-Audio $\times$ CREMA-D & rms\_std (0.69) & rms\_mean (0.18) & rms\_std (0.63) & rms\_mean (0.06) \\
Qwen2.5-Omni $\times$ CREMA-D & rms\_std (0.62) & rms\_mean (0.16) & rms\_std (0.62) & rms\_mean (0.09) \\
Audio-Flamingo-3 $\times$ CREMA-D & rms\_std (0.73) & rms\_std (0.31) & rms\_std (0.65) & rms\_std (0.10) \\
Kimi-Audio $\times$ CREMA-D & rms\_mean (0.64) & rms\_mean (0.32) & rms\_std (0.75) & rms\_std (0.37) \\
Phi-4-MM $\times$ CREMA-D & rms\_mean (0.17) & duration (0.12) & rms\_std (0.62) & rms\_mean (0.17) \\
Qwen2-Audio $\times$ VESUS & rms\_std (0.18) & rms\_mean (0.07) & rms\_mean (0.22) & rms\_mean (0.17) \\
Qwen2.5-Omni $\times$ VESUS & rms\_std (0.10) & rms\_std (0.10) & rms\_mean (0.15) & rms\_mean (0.13) \\
Audio-Flamingo-3 $\times$ VESUS & rms\_std (0.35) & rms\_mean (0.22) & rms\_mean (0.36) & rms\_mean (0.32) \\
Kimi-Audio $\times$ VESUS & rms\_mean (0.12) & rms\_mean (0.04) & rms\_mean (0.12) & rms\_mean (0.06) \\
Phi-4-MM $\times$ VESUS & rms\_mean (0.02) & rms\_mean (0.01) & rms\_mean (0.13) & rms\_mean (0.09) \\
\bottomrule
\end{tabular}
\end{table*}

\subsection{Extended Panel and Nonlinear Removal}
\label{app:sperp-ext}

Table~\ref{tab:app-sperp-ext} repeats the residualized decoding of the main text under descriptor removal of increasing strength: the ten-descriptor panel removed linearly, the extended twenty-descriptor panel removed linearly, and the extended panel removed with gradient-boosted trees. Held-out \(S_{\mathrm{decoding}}\) decodability remains above the 0.25 chance level in every condition under every variant. The descriptor-only reference also strengthens slightly with the extended panel, from 0.640 to 0.664 on CREMA-D and from 0.343 to 0.378 on VESUS, confirming that the added features carry usable information that \(S_{\mathrm{decoding}}\) nevertheless exceeds.

\begin{table}[t]
\centering\small
\setlength{\tabcolsep}{3pt}
\caption{Held-out \(S_{\mathrm{decoding}}\) decodability after descriptor removal of increasing strength. resid-10: ordinary least squares on the ten-descriptor panel (main text). resid-20: the same on the extended twenty-descriptor panel. GBRT-20: gradient-boosted-tree removal of the extended panel. Chance is 0.25.}
\label{tab:app-sperp-ext}
\resizebox{\columnwidth}{!}{%
\begin{tabular}{llrrrr}
\toprule
System & Corpus & \(S_{\mathrm{decoding}}\) & resid-10 & resid-20 & GBRT-20 \\
\midrule
Qwen2-Audio  & CREMA-D & 0.949 & 0.856 & 0.832 & 0.752 \\
Qwen2.5-Omni & CREMA-D & 0.858 & 0.655 & 0.630 & 0.562 \\
Audio-Flamingo-3 & CREMA-D & 0.940 & 0.847 & 0.835 & 0.752 \\
Kimi-Audio   & CREMA-D & 0.823 & 0.610 & 0.574 & 0.494 \\
Phi-4-MM     & CREMA-D & 0.668 & 0.413 & 0.385 & 0.347 \\
Qwen2-Audio  & VESUS   & 0.606 & 0.537 & 0.514 & 0.463 \\
Qwen2.5-Omni & VESUS   & 0.561 & 0.493 & 0.453 & 0.410 \\
Audio-Flamingo-3 & VESUS & 0.690 & 0.651 & 0.610 & 0.568 \\
Kimi-Audio   & VESUS   & 0.563 & 0.488 & 0.474 & 0.424 \\
Phi-4-MM     & VESUS   & 0.481 & 0.423 & 0.380 & 0.349 \\
\bottomrule
\end{tabular}
}
\end{table}

\subsection{Input-Side Loudness Equalization}
\label{app:loudnorm}

The residualization analyses remove surface cues from the state side. The input-side check removes absolute level from the audio itself: every clip is RMS-equalized to a fixed target with peak limiting, answer-position states are re-extracted under the main protocol (the same clips, the four multiple-choice prompts, and the per-condition \(L^*\)), and the subspace decoding is repeated. Two arms are scored. The replication arm rebuilds \(S_{\mathrm{decoding}}\) and refits the probe on equalized training-speaker rows with the split held fixed; \(V_3\) is a function of the model weights and is reused unchanged. The transfer arm applies the probes fit on raw states, without refitting, to the equalized held-out rows.

\begin{table*}[t]
\centering\small
\setlength{\tabcolsep}{4pt}
\caption{Raw versus loudness-equalized held-out decodability under the main protocol. eq: \(S_{\mathrm{decoding}}\) and probe rebuilt on equalized audio with the split held fixed. transfer: raw-fit probe applied without refitting to equalized held-out clips. The last column is the paired raw-minus-equalized \(S_{\mathrm{decoding}}\) difference with speaker-clustered bootstrap 95\% intervals.}
\label{tab:app-loudnorm-np}
\begin{tabular}{llrrrrrl}
\toprule
System & Corpus & \(V_3\) raw & \(V_3\) eq & \(S_{\mathrm{decoding}}\) raw & \(S_{\mathrm{decoding}}\) eq & \(S_{\mathrm{decoding}}\) transfer & \(S_{\mathrm{decoding}}\) raw\(-\)eq [95\% CI] \\
\midrule
Qwen2-Audio  & CREMA-D & 0.890 & 0.891 & 0.949 & 0.949 & 0.950 & $+0.000$\,[$-0.006,+0.006$] \\
Qwen2.5-Omni & CREMA-D & 0.732 & 0.710 & 0.858 & 0.857 & 0.843 & $+0.001$\,[$-0.008,+0.009$] \\
Audio-Flamingo-3 & CREMA-D & 0.937 & 0.933 & 0.940 & 0.939 & 0.939 & $+0.001$\,[$-0.004,+0.007$] \\
Kimi-Audio   & CREMA-D & 0.602 & 0.589 & 0.823 & 0.802 & 0.793 & $+0.022$\,[$+0.008,+0.036$] \\
Phi-4-MM     & CREMA-D & 0.342 & 0.329 & 0.668 & 0.655 & 0.645 & $+0.013$\,[$-0.002,+0.028$] \\
Qwen2-Audio  & VESUS   & 0.463 & 0.460 & 0.606 & 0.584 & 0.599 & $+0.022$\,[$-0.002,+0.047$] \\
Qwen2.5-Omni & VESUS   & 0.464 & 0.448 & 0.561 & 0.544 & 0.551 & $+0.016$\,[$+0.001,+0.036$] \\
Audio-Flamingo-3 & VESUS & 0.670 & 0.649 & 0.690 & 0.694 & 0.673 & $-0.004$\,[$-0.013,+0.006$] \\
Kimi-Audio   & VESUS   & 0.408 & 0.401 & 0.563 & 0.570 & 0.563 & $-0.007$\,[$-0.014,+0.001$] \\
Phi-4-MM     & VESUS   & 0.259 & 0.258 & 0.481 & 0.475 & 0.472 & $+0.007$\,[$-0.013,+0.028$] \\
\bottomrule
\end{tabular}
\end{table*}

Table~\ref{tab:app-loudnorm-np} shows both arms. Held-out \(S_{\mathrm{decoding}}\) accuracy changes by at most 0.022 under replication and 0.030 under transfer across all ten conditions, and the paired raw-minus-equalized interval includes zero in eight of ten. Absolute recording level therefore does not explain most of the readout-external decodability. This control removes only absolute level, not energy dynamics, fundamental frequency, or spectral cues; gain normalization in the audio front ends may also contribute to the observed robustness.

\ifdefined\supplyincludedinpreprint
\else
  \bibliographystyle{IEEEtran}
  \bibliography{IEEEabrv,custom}
\fi
\supplyenddocument

\fi
 
\end{document}